\documentclass[english]{article} % `english' is necessary for babel, not `letter' or `a4paper'
\usepackage[T1]{fontenc}
\usepackage[latin9]{inputenc}\usepackage{imakeidx}
\makeindex
\usepackage{geometry}
\usepackage{fancyhdr}
\usepackage{color}
\definecolor{mypurple}{RGB}{128, 0, 128}
\definecolor{myTeal}{RGB}{0, 128, 128}
\definecolor{myDarkGreen}{RGB}{0, 92, 0}
\definecolor{myLightGreen}{RGB}{0, 192, 0}
\definecolor{myDarkRed}{RGB}{192, 0, 0}
\definecolor{myDarkBlue}{RGB}{0, 0, 192}
\definecolor{schrift}{RGB}{0,73,114}
\usepackage{amsmath} %% get AMS standards for theorems, e.g. the hollow QED box,  \qedsymbol
\usepackage{amssymb} % \sum

\usepackage{hyperref}   % clickable links into or outfrom this text
\hypersetup{
    colorlinks=true,
    linkcolor=myDarkBlue,
    filecolor=magenta,      
    urlcolor=cyan,
    citecolor=myTeal
}

\numberwithin{equation}{section} % number equations as (sec.eq), not just (eq)

\usepackage{graphicx}
\usepackage{sectsty}

\sectionfont{\color{schrift}}  % sets colour of sections
\subsectionfont{\color{schrift}}
\subsubsectionfont{\color{schrift}}

\usepackage{lastpage} 
\usepackage{babel}
\begin{document}
\begin{titlepage}

\title{Elo Ratings for Large Tournaments of \linebreak Software Agents in Asymmetric Games\footnote{
This work was partially sponsored by the Defense Advanced Research Project Agency under award number HR001118C0145. The views, opinions, and/or findings expressed are those of the authors and should not be interpreted as representing the official views or policies of the Department of Defense or the U.S. Government.}}
\author{Ben P. Wise\footnote{Group W, email: bwise@groupw.com.}}

\date{}

\maketitle

\begin{abstract}
The Elo rating system has been used world wide for individual sports and team sports,
as exemplified by the European Go Federation (EGF), International Chess Federation (FIDE) ,
 International Federation of Association Football (FIFA), and many others. To evaluate the
 performance of artificial intelligence agents, it is natural to evaluate them
 on the same Elo scale as humans, such as the rating of 5185
 attributed to AlphaGo Zero \cite{MasteringGo2017}
\\ \\ 
There are 
several fundamental differences between humans and AI that suggest 
modifications to the system, which in turn require revisiting Elo's fundamental rationale. 
AI is typically trained on many more games than humans play, and we have
little a-priori information on newly created AI agents.
Further, AI is being extended into games which are asymmetric between the players, and which
could even have  large complex boards with different setup in every game, such as
commercial paper strategy games.
We present a revised rating
system, and guidelines for tournaments,  to reflect these  differences.
\\ \\
\noindent\textbf{Keywords:} machine learning, 
artificial intelligence,
asymmetric games, 
strategy, games,
chess, performance, rating, Elo, FIDE, FIFA, playing strength, binomial,
trinomial, statistics, numerical
methods, uncertainty.

\end{abstract}

\end{titlepage}

\tableofcontents
\clearpage
\listoffigures
\clearpage
\listoftables
\clearpage

\raggedright
 
\section{Introduction}
There are 
several fundamental difference between humans and AI, or between Chess and commercial paper strategy games, 
which require  modifications to the system, which in turn require revisiting Elo's fundamental rationale. 
We present a revised rating system, and guidelines for tournaments,  to reflect 
the following fundamental  differences:

\begin{itemize}
\item AI is typically  trained and evaluated on many more games than occur in ordinary Chess tournaments. 
The change in scale causes the classic Elo system to behave poorly.
This is discussed in section \ref{sec:Adjusting_Elo_Ranks}.
  
\item While the variability of human performance has been estimated from the extensive literature on ordinary 
Chess tournaments, the variability of each newly trained AI system is completely unknown. 
This is discussed in section \ref{sec:Adjusting_Variance}.
  
\item The Elo rating system is based on a fairly strong assumption about the ability
to rank players on an ordinal scale. This is known to be inaccurate in important cases,
and it is easy to construct plausible examples which violate this assumption. Relaxing
this assumption leads not only to a more complete inventory of the uncertainty in ratings
but also to a simpler and faster algorithm for updating ratings.
This is discussed in section  \ref{sec:Elo_Assumptions}.
  
\item Chess and Go are symmetric games, while commercial paper strategy games typically are not: 
opposing players typically
have different forces, on different terrain, and judge 
performance by different criteria,  which  change from scenario to scenario. 
Methods to take this into account are discussed in section \ref{sec:Asymmetric-Games-Rating-ANOVA}.
 
\item In most strategy games,  succeeding with a comfortable margin is better than 
with a narrow one. In Chess tournaments, outcomes are scored merely 
as win, lose, or tie, with no estimated margin of victory. Methods
to take account of the margin of victory are discussed in section \ref{sec:Margin_of_Victory}.
\end{itemize}

\clearpage

\section{The Classic Elo Rating System}

Arpad Elo\footnote{The Hungarian pronunciation of the last name 
is more like ``ee-loo'' than ``ee-low''.} developed his rating system initially for the game of Chess;
it was adopted by the US Chess 
Federation in 1960 and the International Chess Federation (FIDE) in 1970.
As of September 2020, the US Chess 
Federation used the Elo formula to define the ``Standard Winning Expectancy''
at the core of their rating system.

We will frequently cite Chess as a running example, but the Elo system has been used not only in Chess 
but also in Go, Scrabble, Diplomacy, and many other board games.
The European Go Federation (EGF) uses a system that incrementally updates
players' ratings after each tournament. 
It has also been used in one-on-one 
non-board games, such as tennis and video games. Similarly, it has been used to rate teams in 
American football, European football, basketball, baseball, and others. The official FIFA World
Rankings have used a version of the Elo system since 2018. As will become clear from 
the fundamentals discussed below, it can be adapted to any situation in which a meaningful 
win-or-lose comparison can be made.

\subsection{Bradley-Terry Model}

The Bradley-Terry model is used to estimate the relative strength of two players. It forms the 
foundation of the rating systems for individual sports like Chess, Go, or Tennis, as well as team sports
 such as American or  European Football  \cite{RankingSystemsInFootball}.
 It simply states that the probability one will defeat the other 
 depends on the ratio of their strengths. Note that this is taken as the operational definition of
 ``strength'', as estimated from observed probabilities of victory: 
strength is whatever increases the odds of success. This operational definition
has been successfully extended to political and economic strength of contending
factions and groups
\cite{efird2015toward}
 \cite{Napoli2018}
\cite{Sublinear16}
\cite{IntroKTAB15}
\cite{MultiDimKTAB15}.

For reasons which will become clear later, we carefully distinguish between a ``contest'' and a ``comparison''.
In Chess or Go, each comparison in the Bradley-Terry model corresponds directly to exactly one  contest.
For more general games, it may be necessary to rearrange the results from contests to make sure
that only fair comparisons are made.
 
\begin{equation}
P[A \succ B] =   \frac{S_A}{S_A + S_B}  
\label{eq:Bradley_Terry}
\end{equation}

Thus, if actor $A$ is observed to defeat $B$ about 50\% of the time, their strengths are inferred to 
be equal. If $A$ wins 75\% of the time, then $S_A$ must be three times $S_B$.  If we rate players
 from weakest to strongest, where each has a constant probability of defeating the earlier one
  (e.g. 75\%) then their strengths will be exponentially increasing (e.g. 1, 3, 9, 27, and so on). 
  To avoid inconvenient large numbers, it is common to define a ``rating'' as the logarithm of the
   strength, so that the ratings would increase linearly. 

Note that we can swap $A$ and $B$ to calculate $P[B \succ A]$ and observe that the two 
probabilities add up to exactly 1, not more or less. This means that the Bradley-Terry model
requires that either $A$ wins or $B$ wins: it is not possible for both to win or for both 
to lose or a draw to occur. Nevertheless, it is common practice to use a floating point
margin of victory in the Elo system, which is  1.0, 0.5, or 0.0 for a win, draw, or loss;
see section \ref{sec:Margin_of_Victory}.

\begin{equation}
S =   e^{\beta R} 
\label{eq:Generic_Rank_Definition}
\end{equation}

The constant $\beta$ can be chosen for convenience. In Elo's system, 
equation   \eqref{eq:Elo_Rank_Definition}, $\beta = \ln(10)/400$, as follows:
 
\begin{equation}
S=   10^{R/400} 
\label{eq:Elo_Rank_Definition}
\end{equation}

To avoid the inconvenience and round-off error of handling large $S$ values that largely cancel, equations
\eqref{eq:Bradley_Terry} and 
 \eqref{eq:Elo_Rank_Definition}
 are often combined in the Elo system to get the following:

\begin{eqnarray}
P[A \succ B] & = &   \frac{S_A}{S_A + S_B}   \nonumber \\
  & = &  \frac{e^{\beta R_A} }{e^{\beta R_A} + e^{\beta R_B} }   \nonumber \\
  & = &  \frac{1 }{1 + e^{\beta (R_B - R_A)} }   \nonumber \\
  & = &  \frac{1 }{1 + 10^{ (R_B - R_A)/400} }  
\label{eq:Elo_Prob_Formula}
\end{eqnarray}

Thus, equal Elo ratings correspond to 50/50 odds, while a 200 point advantage in Elo rating
 corresponds to about 75\% chance of victory\footnote{A 75\% probability of
victory actually corresponds to an advantage of 190.85 points, while a 200 point
advantage gives 75.97\% chance of victory. As even a 20 point advantage
would take hundreds of games to reliably detect, differences on the order
of a dozen Elo points have little practical significance.}. Of course, Elo chose $\beta$ so as to 
 approximate the easily-remembered numbers of 200 and 75\%.

Note that in the Elo system, only the difference in ratings matters: if all ratings were raised by 
500 points, or all decreased by 500 points, it would not matter. 
In other words, there is no intrinsic  way to determine a ``zero point''. It is defined by
a simple convention:  in chess, human beginners are assigned an initial
rating of 1000, and the minimum possible rating is 100. Thus, the set of absolute beginners has a fixed, unalterable rating of 1000, 
and individual ratings are adjusted as they accumulate a record of play (and thus are not ``absolute beginners'' any more).
It has been suggested that the minimum rating of 100 was introduced to allow the same rating system
to be used for humans and for software when some programs play almost randomly.

\subsection{Adjusting Elo Ratings}
\label{sec:Adjusting_Elo_Ranks}

Equation \eqref{eq:Elo_Prob_Formula} is usually cited as the foundation of the Elo system, and many ingenious 
algorithms have been devised to estimate strengths from an entire database of win/loss
records.\footnote{Our proposed approaches  to using an entire database
can be found in sections \ref{sec:Batch_Adjust} and  \ref{sec:Elo_Assumptions}.}
However, when new games are played, how can the new win/loss  be incrementally combined with the 
old records to get new estimates, without reanalyzing the entire database? 
This is necessary so that a simple hand-calculation can be done, on-site after a tournament, 
using only the data from that tournament and not the complete database of all tournaments.
Humans constantly change as they learn and age, so incremental update is clearly 
necessary for people on the scale of years. However, they do not change significantly
during the span of a single tournament. Therefore, a system was needed that
met three criteria:

\begin{itemize}
	\item Feasible to calculate  with the hand-held calculators of Elo's day
	\item Process all the results of a tournament in one batch, simultaneously
	\item Incrementally update performance across tournaments.
\end{itemize}

Elo's formulation was to treat the assigned rating as an estimate of the true rating based on prior tournaments,
then apply 
Bayes' Theorem to update that estimate using all the games from the current tournament.  
Elo assumed that the errors in measuring a particular player's rating are normally distributed
around that player's true rating (not that true ratings are normally distributed). 
A  normal distribution is characterized 
by its mean and variance, $\mu$ and $\sigma^2$,
where the prior probability of any given $R$ value given  $\mu$ and  $\sigma$ is as follows:

\begin{equation}
p[R] = \frac {e^{-\frac {(R-\mu)^2}{2 \sigma^2}}} {\sqrt{2 \pi} \ \sigma}
\label{eq:Normal_Distrib}
\end{equation}

Elo formulated the problem as calculating the most likely value of $R$, given the prior 
estimate $\mu$ and the new win/loss records. We will call win/loss records from a
 tournament collectively as $T$, with each individual outcome designated $T_i$. 
 For now, we assume there are $N$ comparisons with $J$ wins and $K$ losses.
 Again, in chess, each comparison refers to a single contest.

We start with Bayes' Theorem for posterior probabilities:
\begin{equation}
p[ R | T ]   =  \frac{ P[ T |R  ]  \ p[R ]  }
                                           {\int_{\rho}  \ P[ T | \rho ] \ p[ \rho ] \ d\rho  } 
\end{equation}

Notice that the denominator is a constant unaffected by $\mu$ or by $\sigma$, so it can be safely ignored
in what follows (because it is a constant, its derivative is zero).

\begin{eqnarray}
P[ T | R  ]  p[R ] & = & \Pi_i^N P[ T_i | \mu, \sigma  ] \  p[ R ]   \nonumber   \\
P & = &  \left( \Pi_{j=1}^{J} P[ R \succ R_j | \mu, \sigma  ]\right) 
                  \    \left( \Pi_{k=1}^{K} P[ R_k \succ R | \mu, \sigma  ]\right) \  p[ R ]   \nonumber
\label{eq:Elo_Posterior_Prob_Formula}
\end{eqnarray}

Because  individual comparisons in the tournament are conditionally independent, 
we can factor the numerator into  $N=J+K$ terms
for $J$ individual wins 
and $K$ individual losses as follows:\footnote{Strictly speaking, the term $R_k$ should be $R_{J+k}$, 
but we simplify notation without ambiguity.}

\begin{eqnarray}
P & = &   \left( \Pi_{j=1}^{J} \frac{e^{\beta R}} {e^{\beta R} + e^{\beta R_j} } \right) 
        \   \left(\Pi_{k=1}^{K} \frac{e^{\beta R_k}} {e^{\beta R} + e^{\beta R_k} }\right) 
        \   \frac {e^{-\frac {(R-\mu)^2}{\sigma^2}}} {\sqrt{2 \pi} \ \sigma}  \nonumber  \\
& = &   \left( \Pi_{j=1}^{J} \frac{1} {1 + e^{\beta (R_j - R)} } \right) 
        \   \left(\Pi_{k=1}^{K} \frac{1} {1 + e^{\beta (R - R_k)} }\right) 
        \   \frac {e^{-\frac {(R-\mu)^2}{2 \sigma^2}}} {\sqrt{2 \pi} \ \sigma}  
\label{eq:Elo_Posterior_Likelihood_Formula}
\end{eqnarray}

Elo's formula is simply the $R$ value to maximize the posterior probability in 
equation \eqref{eq:Elo_Posterior_Likelihood_Formula}.\footnote{The alert reader
will notice a problem: the ratings of opponents are assumed to be known with 
certainty. The implications of this assumption are discussed around
Table  \ref{table:Elo_Overshoot} on page  \pageref{table:Elo_Overshoot}.}
To obtain the simple update formula,
we take the logarithm of both sides, then take the derivative with respect to $R$, 
set it to zero, and solve.
\begin{eqnarray}
\ln(P)
& = &    \sum_{j=1}^{J}  -\ln(1 + e^{\beta (R_j - R)} )  
        \  +  \sum_{k=1}^{K}  -\ln(1 + e^{\beta (R - R_k)} )  
        \  - \frac {(R-\mu)^2}{2\sigma^2} - \ln(\sqrt{2 \pi} \ \sigma)   \nonumber   \\
\frac{\partial \ln(P)} {\partial R} 
& = & 
           \sum_{j=1}^{J}  \beta \frac{e^{R_j}}{e^{R} + e^{R_j} }  
         \ + \sum_{k=1}^{K}  -\beta \frac{e^{R}}{e^{R} + e^{R_k} }  
         \ - \frac{R - \mu}{\sigma^2}  \nonumber   \\
& = & 
         \beta \left(  \sum_{j=1}^{J}  \frac{e^{R_j}}{e^{R} + e^{R_j} }  
         \ - \sum_{k=1}^{K}  \frac{e^{R}}{e^{R} + e^{R_k} }  \right)
         \ - \frac{R - \mu}{\sigma^2} \nonumber   \\
0 & = & 
         \beta \left(  \sum_{j=1}^{J} \left( 1-  \frac{e^{R}}{e^{R} + e^{R_j} }  \right)
         \ - \sum_{k=1}^{K}  \frac{e^{R}}{e^{R} + e^{R_k} }  \right)
         \ - \frac{R - \mu}{\sigma^2}
\end{eqnarray}

This complicated expression simplifies remarkably. First, we notice that both summations 
include the estimated probability that this 
actor would beat the opponent, using the new rating. Obviously, if they have 100\% 
estimated probability of winning each comparison, then
the sums would be  $J$ and $K$ respectively, so we call the sum of the two summations
 simply the expected number of wins, given the new rating. Call that $E(R)$.
The summation over $j$ has 1 in each of $J$ terms, which add up to the number of actual wins, $A = J$. To be specific, if we define a win (by any amount) as $w=1$, a loss (by any amount) as $w=0$, and a draw as $w=0.5$, then A is the sum of the $w$ values.

With these two substitutions, the final line becomes the following
\begin{eqnarray}
0 & = & \beta \left( A - E(R) \right) - \frac{R - \mu}{\sigma^2} 
\end{eqnarray}

Rearranging, we get equation \eqref{eq:Consistent_Elo_Update}, which we call the Self-Consistent Elo (SC-Elo) update,
which is the posterior maximum likelihood (PML) estimate:
\begin{eqnarray}
R & = & \mu + \beta \sigma^2 \left( A - E(R) \right)  
\label{eq:Consistent_Elo_Update}
\end{eqnarray}

One interpretation of \eqref{eq:Consistent_Elo_Update} is that it adds a correction term to
$\mu$. Whenever the actual result is above the expected result, there is a positive correction.
This is similar to the goal-seeking proportional feedback in section \ref{sec:moving_averages}.

The classic Elo update formula is almost the same, but with one crucial difference:
\begin{eqnarray}
R & = & \mu + \beta \sigma^2 \left( A - E(\mu) \right)  
\label{eq:Classic_Elo_Update}
\end{eqnarray}

In the Elo equation, the old estimate of the rating ($\mu$) is used to calculate the estimated 
number of wins, even though
the derivation actually shows that the new estimate of rating ($R$) should be used. 
In other words, the Elo formula relies on the approximation that $E(R) \approx  E(\mu)$,
which  has at least two significant effects.
The first effect  is that the Elo formula has consistent and predictable errors:
it will always adjust by more than the true PML estimate, so 
it is not an ``unbiased estimator'' of the true rating. The problem of overshoot 
with small samples can be explained as follows.
When $A > E$ and the score should be increased, $E$ should be increased from  $E(\mu)$ to $E(R)$, 
reducing the adjustment, but Elo does not do this.
When the score should be increased, Elo always increases it too much, in every case,
even for small tournaments.
When it should be decreased, Elo always decreases it too much.
The second effect is that the adjustment increases linearly with $N$, leading
to implausibly large adjustment when $N$ is large. In ordinary 
human-scale tournaments, which might
involve tens of games by each player at most, the approximation makes little difference. 
In AI training, where there might be several hundred games in a tournament,
the effect can be quite large.
During the training of Alpha Go, each new AI was
evaluated against the previous best in a tournament of 400 games; a 55\% win
 rate (or 34.86 Elo points) was required to accept
the AI as the new best. 
To illustrate the overshoot problem, we
applied equation \eqref{eq:Classic_Elo_Update} for
various $N$, assuming $\mu = 1250$, opponent rating also 1250, $\beta \sigma^2 = 116$, and a 
65\% win rate.
The second column of table  \ref{table:Elo_Overshoot} indicates that Elo cannot be directly applied to a 400-game data set
without risking substantial error. The third column shows that the SC-Elo formula does
not have increasing error.

Over time, over multiple small tournaments, 
the Elo formula should  converge like a feedback control system to the
approximate true rating\footnote{Over multiple large tournaments, the overshoot could cause diverging oscillations,
as is well-known from control theory.} and then do a random walk 
around the true value depending on the
player's luck. However, as we shall see, the consistent over-adjustments are just two
of the reasons why the variance of the Elo estimate (i.e. average error) is much
 higher than that of the SC-Elo.

\begin{table} 
\begin{center}
\begin{tabular} {| r | r r |   }
\hline
N & Elo & SC-Elo   \\
\hline
4       & 1319.6 & 1291.8 \\
40     & 1946.0 & 1342.5  \\
400   & 8210.0 & 1355.8  \\
4000 & 70850.0 & 1357.4   \\
\hline
\end{tabular}
\caption{Overshoot in Classic Elo}
\label{table:Elo_Overshoot}
\end{center}
\end{table}

Large tournaments   present a dilemma.
First, a large tournament of 400 games could be broken up into (for example) forty mini-tournaments 
of ten games each, and the  Elo formula applied forty times in order. 
Obviously, processing 400 games in blocks of 10 each means
that the random variation does not "average out" as much in
10 games as in 400.
As mentioned above, the  Elo estimate
is accurate when averaged over time for non-learning players, but the standard error is quite large due to both overshoot
and lack of smoothing, depending on the players
 luck in each mini-tournament. Second, we could average out
the random variations by processing the whole tournament in one 400-game batch with 
classic Elo - but the results are usually far outside 
the bounds of plausibility.

The problem of over-adjustment in large tournaments can be explained as follows. 
Suppose that the tournament has $N$ games, the estimated probability of beating each
opponent is $\hat{p} = E(\mu) / N$, and  the observed empirical frequency of winning is $f = A/N$. 
We can then write the classic Elo update formula as follows:

\begin{eqnarray}
R & = & \mu + \beta \sigma^2 \left( A - E(\mu) \right)   \nonumber \\
  & = & \mu + \beta \sigma^2 N \left( f - \hat{p} \right) 
\label{eq:Classic_Elo_Update_N}
\end{eqnarray}

As $N$ becomes large, the adjustment in $R$ increases without limit, causing over extrapolation.
 The difference between the prior estimated wins, $E(\mu)$ and the posterior estimate
of wins, $E(R)$, is only a small error for small $N$, but it grows steadily.
The error rapidly becomes quite large: for just a few hundred games, this can cause the 
C-Elo formula to be off by thousands of points. Table  \ref{table:Elo_Overshoot}
shows the results with  65\% win rate, which corresponds to an Elo advantage of $107.54$ points.
As the opponent has rating 1250, the SC-Elo rating approaches $1357.54 = 1250+107.54$
as the ever-larger tournament data outweighs the prior estimate more and more.

\subsubsection{Numerical Stability of SC-Elo}

While equation \eqref{eq:Consistent_Elo_Update} defines the self-consistent estimate, it is not easy to solve directly. 
The basic problem is that the slope of $E(R)$ with respect to $R$
is proportional to the tournament size $N$, so it becomes quite numerically sensitive.
It can be solved with a simple and efficient iterative procedure by explicitly taking the slope into account.

If we designate the estimated probability of victory for an $r$-rated player over opponent $i$ as $p_i (r)$, then we 
can calculate two very useful expressions as follows.

\begin{equation}
E(r)  =  \sum_i  \ p_i (r) 
 \label{eq:SC_Elo_Height}
\end{equation}

\begin{equation}
\frac{\partial E}{\partial r}   =  \beta \ \sum_i  \ p_i (r)  (1-  p_i (r))  
\label{eq:SC_Elo_Slope}
\end{equation}

Note that because only the difference in ratings matters, and the expected Blue wins plus expected Red wins is always $N$,
we have

\begin{equation}
\frac{\partial E_B}{\partial r_B}   =  \frac{\partial E_R}{\partial r_R}  
\end{equation}

Define $s$ as the slope of $E$ at the old estimate $\mu$, and we can approximate $E(R)$ with the 
initial terms of a standard Taylor series expansion:
\begin{eqnarray}
E(R) & \approx & E(\mu) + s \left( R - \mu \right)
\label{eq:Elo_Slope}
\end{eqnarray}

If we use this expression for $E(R)$ in equation \eqref{eq:Consistent_Elo_Update}, we get the following:
\begin{eqnarray}
R & = & \mu + \beta \sigma^2 \left( A - \left(  E(\mu) + s \left( R - \mu \right) \right) \right)    \nonumber \\
 & = & \frac
 {\mu + \beta \sigma^2 \left( A - E(\mu) \right) \ + \  \beta \sigma^2 s \mu}
 {1  \ + \  \beta \sigma^2 s}
\label{eq:Consistent_Elo_Update_Iteration}
\end{eqnarray}

Notice that if $s = 0$, this reduces to the classic Elo formula, because equation
\eqref{eq:Elo_Slope} then reduces to the classic Elo-approximation  $E(R) \approx  E(\mu)$.

We can write these equations in iterative form, starting with $R_0 = \mu$. At any given iteration $t+1$,
equation  \eqref{eq:Consistent_Elo_Update} applies, which gives the first line. We then substitute
the most recent estimate using the most recent slope, and expand terms.

\begin{eqnarray}
R_{t+1} &=& R_0 + \beta \sigma^2 \left( A - E(R_{t+1}) \right)   \nonumber  \\
             &=& R_0 + \beta \sigma^2 \left( A -\left[ E(R_{t}) + s_t (R_{t+1} - R_t ) \right] \right)  \nonumber  \\
             &=& R_0 + \beta \sigma^2 \left( A - E(R_t) - s_t R_{t+1} + s_t R_t  \right)  \nonumber  \\
             &=& R_0 + s_t  \beta \sigma^2 R_t +  \beta \sigma^2 \left( A - E(R_t) \right) - s_t   \beta \sigma^2 R_{t+1}  
\end{eqnarray}

Solving for $R_{t+1}$, we can see how the new estimate is a weighted average:

\begin{equation}
\hat{R}_{t+1}   = 
 \frac
 {R_0  \ + \  \beta \sigma^2 \left( A - E(R_{t}) \right) \ + \ s_{t} \beta \sigma^2  R_{t}  }
 {1  \ + \   s_{t} \beta \sigma^2 } 
\label{eq:Consistent_Elo_Update_Algorithm_first}
\end{equation}

Equation \eqref{eq:Consistent_Elo_Update_Algorithm_first} oscillates around the final value,
so we can significantly improve stability and reduce convergence time by averaging the new estimate
with the most recent value. Any weight between 0 and 1 could be used; we use 0.5 for simplicity.
If we  re-arrange terms to reduce computation, we get  the following,
where $K$ is a constant for this particular update of this particular player:

\begin{eqnarray}
K &=& \beta \sigma^2 \\
 s_{t} &= & \frac{\partial E}{\partial r}\Bigr|_{\substack{R_{t}}} \\
\hat{R}_{t+1}  & = & \frac
 {R_0  \ +  \ K \left( A - E(R_{t}) + s_t R_t  \right) }
 {1  \ + \   s_{t} K }  \\
R_{t+1} &=& \frac{ \hat{R}_{t+1} + R_t} {2}
\label{eq:Consistent_Elo_Update_Algorithm}
\end{eqnarray}

Even for a tournament of several hundred games, equation \eqref{eq:Consistent_Elo_Update_Algorithm} 
generally falls within 10 or so points of the exact
answer on the first iteration and converges to a fraction of an Elo point in three or four iterations.
See section \ref{sec:Precision} for a discussion of how tight a convergence is desirable.

In summary, the  classic Elo system will always severely overshoot in large tournaments, 
producing estimated ratings which are off by
thousands of Elo points: the SC-Elo system should always be preferred for  tournaments of more than a few dozen
games per player.

\subsubsection{Draws}
\label{sec:Draws}

In most board games, it is possible for a draw to occur. Strictly speaking,
the Bradley-Terry and Elo models assume a binary win or lose outcome, not a
ternary win, draw, or lose outcome. In terms of betting on outcomes, this
could matter a great deal when a substantial fraction of grandmaster chess games
result in draws. However, for ratings, the common approach is to assign
a floating-point margin of victory. Each player gets 1.0 for
a win, 0.0 for a loss, and 0.5 for a draw  in the $A$ term of the Elo formula.
This is consistent with $P[A \succ B] = 1/2$ when their ratings are equal. If the ratings are equal 
and they draw every game, then $A = N/2 = E$,
the results were exactly as expected, and the estimated rating is unchanged.

When the actual results are close to (far from) the expected, the uncertainty
in $R$ should decrease (increase) because the evidence confirms (contradicts)
the prior expectation; see section \ref{sec:Adjusting_Variance}

Note that when margin of victory is taken into  account, 
e.g. by equation (\ref{eq:Modified_Final_Reward}),
the actual score $A$ is the sum of the weighted victories:
\begin{equation}
A = \sum_i w_i
\end{equation}

\subsubsection{SC-Elo with uninformative priors}

In the standard usage of Elo ratings, not only do the individual players have past history
but the rating organizations have a great deal of prior experience with ratings. 
Hence, there are widely accepted ways to estimate $\sigma^2$ or $K$ for a player based on
past experience with this or with other players.

However, neither is true for a newly created AI agent.
A standard approach is to use a so-called uninformative prior, which is equivalent to taking the
limit as   $\sigma  \rightarrow \infty$ (see section   \ref{sec:Batch_Adjust} for a numerical example).
This makes $K$ extremely large in
equation \eqref{eq:Consistent_Elo_Update_Algorithm_first}.

\begin{equation}
R_{t+1}  =  R_t  +   \frac {  A - E(R_{t})   } {   s_{t}  }   
\end{equation}

Unsuprisingly, this is the iterative algorithm to reproduce exactly the observed win-fraction:
it converges to $R_{t+1}  =  R_t $ when  $A = E(R)$.

\subsubsection{SC-Elo in Limit of Large Tournaments}
\label{sec:Large_Tournament_Limit}

One of the motivations of \eqref{eq:Consistent_Elo_Update_Iteration} was to deal with 
large tournaments, so we explicitly explore that limit by looking at arbitrarily large tournaments.
This shows that the SC-Elo never overshoots but smoothly approaches an obviously plausible value.
Separating the terms of equation \eqref{eq:Consistent_Elo_Update_Iteration},

\begin{eqnarray}
R & = &  \frac
 {\mu }
 {1  \ + \  \beta \sigma^2 s} 
 \ + \
  \frac
 {  \beta \sigma^2 \left( A - E(\mu) \right)  }
 {1  \ + \  \beta \sigma^2 s} 
 \ + \
  \frac
 {  \beta \sigma^2 s \mu}
 {1  \ + \  \beta \sigma^2 s} 
\end{eqnarray}

As the tournament size $N$ increases, $s$ increases proportionally, so we can write the
 limit as $N \rightarrow \infty$ as

\begin{eqnarray}
R & = & 0  \ + \ 
  \frac
 {    A - E(\mu)  }
 {  s}
 \ + \  \mu 
\end{eqnarray}

Because $s = \partial E / \partial R$, this turns into another Taylor expansion:

\begin{eqnarray}
R & = \mu +  \frac{\partial R}{\partial E}
\ \left( A - E(\mu)  \right) 
\label{eq:scelo_large_limit}
\end{eqnarray}

This shows that even for arbitrarily large tournaments, the updated rating $R$ converges to a definite limit,
which  is precisely the value  needed to  reproduce exactly the observed win-fraction.
Notice that the initial uncertainty $\sigma$ in the rating does not appear in the large-tournament estimate. 
This is because the large amount of 
empirical data from a large tournament completely outweighs any prior information.

While equations \eqref{eq:Consistent_Elo_Update} and \eqref{eq:Consistent_Elo_Update_Algorithm} 
solve the problem of updating
ratings in both large and small tournaments, they do not address one crucial question: how do we 
estimate $\sigma$?

\subsection{Adjusting Variances}
\label{sec:Adjusting_Variance}

While equation \eqref{eq:Classic_Elo_Update} defines the Elo update,  it is usually written in 
a different form, with an empirical parameter $K$:

\begin{eqnarray}
R' & = & R + K \left( A - E \right)
\end{eqnarray}

By comparison of formulas, it is clear that $K = \beta \sigma^2$. In the Elo system for Chess, the $K$ values were
at one point set  
to  $116$ for beginners, $10$ for very skilled players, and 32 for everyone in between.
The logic is directly tied to empirical observations about $\sigma$. For beginners, there is simply a wide range of talent, 
but no record of performance. Further, they are likely to learn most quickly, so even
if the rating were known precisely last year, it could easily have changed significantly over a year. For grand masters,
 there is obviously a long record of games on which to base the rating (which reduces
uncertainty) and they are likely to improve their skills much more slowly.  Given that $\beta = \ln(10) / 400$, this means
 that the empirical estimates for human players are $\sigma$ of $142.0$, $74.6$, and $41.7$
respectively. Note that an Elo-advantage of $42$ points gives only estimated odds of $56\!:\!44$, so 
uncertainty of $\pm 41.7$ is plausibly small.

As discussed in section \ref{sec:Total_of_Ranks}, having lower $K$ values for players with higher ratings
turns out to be crucial to the fairness and stability of the whole system. Therefore, for  any of the methods
proposed here, it must be verified that the necessary statistical  condition applies to the system as a whole.

When developing a series of AI players, we cannot simply assume that they have the same variability as do humans.
It might be that a newly developed AI agent plays extremely consistently (consistently well or consistently poorly), and so 
had a very low $\sigma$. 
It might be  that the agent is extremely good in some areas, but very weak in others, giving very
 inconsistent play and hence high $\sigma$. For people, the most highly rated players have long records
of play, so it is intuitively plausible that their $\sigma$ (i.e. $K$) values should be low. However, an
AI agent might be trained on a large data set before its first Elo rating, so it might produce
grand master performance in its very first rated game: it would be manifestly wrong to assign
it an initial Elo rating of $1000 \pm 100$  when we already know it was able to consistently defeat AI 
opponents which themselves had 2500 ratings. We can estimate the
average Elo rating of one or several new agents using equations
\eqref{eq:Advantage_from_Log_Odds} and \eqref{eq:Population_Rating_Improvement}.
The $\sigma$ value for an AI agent must be estimated from the observed play of that agent,
and fortunately there are several reasonable candidate methods to do so
in  Elo's framework.  The best estimate is from equation 
\eqref{eq:Nonlinear_Elo_errors}
which also takes into account the degree to which observed data does not
fit Elo's framework.

\subsubsection{Estimation from Series}

The standard way to estimate the mean of a series is to take a simple average.
 If the average might change over time, 
the series is non-stationary, and the optimal estimate is an exponentially
weighted moving average, e.g. Kalman filtering. Of course, a stationary process with stable parameters is just a 
special case of a non-stationary process. 
The variance of any quantity is defined as the average of the squared deviation, so the optimal estimate of the variance
 of a non-stationary process is an exponentially weighted moving average of the square differences.

This leads directly to a simple formula for updating an estimated variance, $\sigma^2$. 
We count tournaments with $t$, where
 the most recent had $N_t$ games. The parameter $M$ reflects a judgment as to 
how many previous games of the AI agent should be used to estimate $\sigma$ (note that this is
not an estimate of $\sigma$ itself).
First, we update the estimated rating from $R_{t-1}$ to $R_t$ using SC-Elo:

\begin{eqnarray}
R_t & = R_{t-1} + \beta \sigma^2 \left( A_t - E(R_t) \right)
\end{eqnarray}

Second, we update the moving average estimate of the variance, depending on how much $R$ changed:
\begin{eqnarray}
\sigma_{t}^2 & = & \frac
{N_t \left(  R_{t} - R_{t-1} \right)^2 + M \sigma_{t-1}^2 }
{N_t + M}
\label{eq:Consistent_Elo_Moving_Average_Uncertainty}
\end{eqnarray}

\subsubsection{Estimation from Binomial Process}

Particularly with large tournaments, it is theoretically possible  to estimate directly how much 
noise is in the observation $A_t$, and
 hence a theoretical lower bound on  $\sigma_t$, the uncertainty in $R$.
The first key observation is that a tournament between an AI agent and some opponents is a 
binomial process, for which the mean
 and variance are well understood.
The second key is to observe that if $Y = cX$, and the uncertainty in $X$ is $\sigma_X$, then $\sigma_Y = c \sigma_X $.
This is like a conversion of units: because one dollar is one hundred cents,
plus or minus one dollar is the same thing as plus or minus  one hundred cents.

If we have $N$ trials of a binomial process that has probability $p$ of success at each trial,
 then the mean number of actual successes is $\mu[A] = Np$ 
 with variance $\sigma^2[A] = N p (1-p)$. Notice that $\sigma[A]$ 
only grows
as $\sqrt{N}$, so the relative error  $\sigma[A] / \mu[A] $  in our empirical estimate of $p$  will fall as $\sqrt{N}$. 
 As the estimated probabilities
in a tournament vary with the opponent, the expected number of actual  successes and the 
variance, given $r$,  
are quite similar to equations \eqref{eq:SC_Elo_Height} and \eqref{eq:SC_Elo_Slope}\footnote{Hence, we can calculate 
$\sigma^2[A | r]$ and $s = \beta  \sigma^2[A | r]$ at each iteration of the SC-Elo 
algorithm for calculating $R$, then use the
 final $\sigma[A|R]$ to calculate the final uncertainty at negligible computational cost.}.

\begin{eqnarray}
\mu[A | r] & = & \sum_i  \ p_i (r)   \\
\sigma^2[A | r]  & = & \sum_i  \ p_i (r)  (1-  p_i (r))
\label{eq:Mean_and_Stdev_Actual}
\end{eqnarray}

We can replace $A$ by $A \pm \sigma[A]$ in equation \eqref{eq:Consistent_Elo_Update_Iteration} to
 immediately get an estimate of the posterior uncertainty in $R$, which we will designate $\sigma_t$.
If we use this in equation \eqref{eq:Consistent_Elo_Update}, we get the following expression for
 the uncertainty in $R$,  where $R_t$ is the SC-Elo update, and $s$ is the slope at $R_t$ (not $\mu$):
\begin{eqnarray}
\sigma_t & = &    \frac  {\beta \  \sigma_{t-1}^2 \sigma[A_t|R_t]}
 {1  \ + \  \beta \ \sigma_{t-1}^2 s} 
\label{eq:Consistent_Elo_Uncertainty} \\
  & \approx &    \frac  {  \sigma[A_t|R_t]}
 {  s}
\label{eq:Consistent_Elo_Uncertainty_Approx}
\end{eqnarray}

Notice  that the uncertainty in $R$ should
fall roughly as $\sqrt{N}$ for large $N$, as expected.
This happens because  $\sigma[A|R]$ grows as $\sqrt{N}$ in the numerator 
while $s$ grows as $N$ in the denominator.
Equation \eqref{eq:Consistent_Elo_Uncertainty} is used in 
Table  \ref{table:Elo_Overshoot} to illustrate the
scale  of uncertainty in $R$ which the binomial approach produces.

Again, we can write out $s$ as a partial derivative and see that equation
\eqref{eq:Consistent_Elo_Uncertainty_Approx}  is actually a simple conversion of units,
 from uncertainty in outcomes ($A$ and $E$) to uncertainty in ratings ($R$).
\begin{eqnarray}
\sigma_t & = &    \frac{\partial R}{\partial E} \  \sigma[A_t|R_t]
\label{eq:Consistent_Elo_Uncertainty_Conversion}
\end{eqnarray}

Because we know that equation  \eqref{eq:Consistent_Elo_Uncertainty} is the 
theoretical lower bound on the uncertainty
in $R$, it is suggested that the estimated variance of an actor's rating be updated
 to the maximum
of equations 
\eqref{eq:Consistent_Elo_Moving_Average_Uncertainty}
and
\eqref{eq:Consistent_Elo_Uncertainty}. The ideal approach would be to compute the posterior likelihood
of $R$, taking into account not only that the uncertainty in rating is normally distributed but 
also that the uncertainty in tournament results is binomially distributed. However, the algebra was found to 
be  intractable.

\subsubsection{Estimation from Posterior Distribution}
\label{sec:Posterior_Variance_Estimation}

The original goal of Elo was to develop a formula  simple enough to apply
at tournaments
with the tools of his day, such as a hand-held calculator.  However, the SC-Elo computation
is complex enough that a general-purpose computer is required: we pay no additional computational
cost by looking at the full Bayesian computation.

Equation  \eqref{eq:Elo_Posterior_Prob_Formula} gives the posterior probability of a rating,
up to a normalizing constant. If we take $p[R] = 1$ as an uninformative prior,
we can numerically calculate the posterior probability (up to a constant) and thus
directly compute the first and second moments of the posterior distribution.

For a set of possible ratings, ${ r_1 , ..., r_n}$, we can use equation  \eqref{eq:Elo_Posterior_Prob_Formula}
to calculate  ${ p_1 , ..., p_n}$ (except for the constant denominator). Then we can calculate
the posterior mean and variance:

\begin{eqnarray}
Q & = &   \sum_i p_i  \\
M_1 (R) &=&  \frac{ \sum_i r_i p_i }{Q} \\
M_2 (R)&=&  \frac{ \sum_i r^2_i p_i }{Q} \\
R_1 &=& M_1(R) \\
\sigma^2 &=& M_2(R) - R^2_1
\label{eq:Posterior_Mean_and_Variance}
\end{eqnarray}

As long as the players have  mixed records of wins and losses, the posterior likelihood function has
a clear peak: the PML estimate is very close to the true Bayesian expectation. The benefit of
equation \eqref{eq:Posterior_Mean_and_Variance} is not so much a new estimate for$R_1$ but
for $\sigma$.

We have found that taking the range of possible ratings to be $n=100$ values,
evenly spaced from 200 points below the
rating of the weakest player to 200 points above the  rating of the strongest player was quite accurate. 
Raising $n$ to 1000 or more typically changed the $R_1$ and $\sigma$ by only a fraction of an Elo
point, which is insignificant when $\sigma$ is on the order of 50 Elo points or so.

Of course, a more theoretically consistent approach would be to postulate prior distributions
over both $R$ and $\sigma$, then do a Markov Chain Monte Carlo calculation to get the posterior
probabilities, and finally take the posterior expectations of $R$ and $\sigma$. However, it
is expected that the difference would be small compared to the uncertainty
arising from the binomial nature of the contests themselves.

\subsection{Elo Averages}

Equation \eqref{eq:Elo_Prob_Formula} can be used not only to rate individual players but also to determine useful summary
statistics about populations. We review a generalized definition of averages before applying it
to define two different kinds of "Elo average."

\subsubsection{Generalized Averages}

 The essence of the averaging operation is to find a single value, $a$, which
gives the same result when it is substituted for a set of $x_i$ values:

\begin{eqnarray} 
\sum_i  f(x_i, w_i) &=& \sum_i  f(a, w_i) 
\label{eq:Generalized_Averages}
\end{eqnarray}

Obviously, the product of  terms is just the sum of their logarithms, so equation 
\eqref{eq:Generalized_Averages} covers both cases, as well as any other nonlinear
transformations of the individual terms.

While the most common kind of average is a simple arithmetic average, there are many useful variations on the concept.
In equation \eqref{eq:Basic_Averages}, the value $a$ is a kind of weighted average of the various $x_i$, given the
vector of parameters $\theta  = [w_1, ..., w_n, d]$ for 
weights $w_i$ and the exponent $d$.

\begin{eqnarray}
f(x,w) &=& x^w \nonumber \\
\sum_i  w_i x^d_i  &=&  \sum_i  w_i a^d 
\label{eq:Basic_Averages}
\end{eqnarray}

If $d=1$, equation \eqref{eq:Basic_Averages} specifies $a$
as the familiar weighted  arithmetic average. When $d=2$, $a$  is the  weighted root mean square.

As $ d \rightarrow 0$, $a$ becomes the  weighted geometric average:
\begin{eqnarray}
\sum_i  w_i x^d_i  &=&  \sum_i  w_i a^d  \nonumber  \\
\sum_i  w_i e^{d \ln(x_i)} &=&  \sum_i  w_i e^{d \ln(a)} 
\end{eqnarray}

Because $e^z \approx 1+z$ for small $z$,

\begin{eqnarray}
\sum_i  w_i (1 + d \ln(x_i)) &=&  \sum_i  w_i  (1 + d \ln(a))   \nonumber \\
\sum_i  w_i d \ln(x_i) &=&  \sum_i  w_i   d \ln(a)   \nonumber \\
\sum_i  w_i   \ln(x_i) &=&  \sum_i  w_i   \ln(a)   \nonumber \\
\Pi_i x_i^{w_i} &=& \Pi_i a^{w_i} 
\end{eqnarray}

The arithmetic, geometric, and RMS averages all share the property that when
one $x_i$ increases without limit, so does $a$: the result is dominated by the largest term.

When $d= -1$, equation \eqref{eq:Basic_Averages} defines the weighted harmonic average:

\begin{eqnarray}
\sum_i { \frac{w_i}{x_i}}  &=&  \sum_i   \frac{w_i}{a} 
\end{eqnarray}

The harmonic average is commonly encountered when considering velocities.
Unlike the other three, the harmonic average is not dominated by the largest term.
For example,
if a particle traverses one mile at one mile per hour, then another mile at infinite speed, then
it takes one hour to go two miles. It goes the same distance in the same
time as if it had moved at a steady pace of two miles per hour. The harmonic average of $1$ and $+\infty$ is $2$:

\begin{equation}
\frac{1}{1} + \frac{1}{\infty} = 1 = \frac{1}{2}  + \frac{1}{2} 
\end{equation}

\subsubsection{Averages of Mixed Capabilities}
\label{sec:Mixed_Capabilities}

A similar effect occurs with equation \eqref{eq:Elo_Prob_Formula}, because the rating-difference
appears in the denominator. Suppose player X has an Elo rating of 1000.
Suppose player Y is a software agent that half the time plays at 1000, but half the time plays to the best
of its ability and defeats X every time, which would have the same effect as an enormous Elo rating. Thus,
Y would win 75\% of the time, which corresponds to an advantage of 191 Elo points. Thus, the "elo-average"
of $0$ points advantage and $\infty$ points advantage is $191$ points.

\begin{table} 
\begin{center}
\begin{tabular} {| c c | c |  }
\hline
 Agent X & Agent Y & $P[Y \succ X]$ \\
\hline
 \color{myDarkRed}1200 &  \color{blue}1400   & 0.7597 \\
 \color{blue}1200 &  \color{myDarkRed}1100   & 0.3599 \\
 1200 & 1241.8   & 0.5599 \\
\hline
\end{tabular}
\caption{Average Ratings with One Mixed Capability}
\label{table:Elo_Average_00}
\end{center}
\end{table}

\begin{table} 
\begin{center}
\begin{tabular} {| c c | c |  }
\hline
 Agent X & Agent Y & $P[Y \succ X]$ \\
\hline
 \color{myDarkRed}1200 &  \color{blue}2500   & 0.9994 \\
 \color{blue}1200 &  \color{myDarkRed}1100   & 0.3599 \\
 1200 & 1330.7   & 0.6797 \\
\hline
\end{tabular}
\caption{Average Ratings with High Blue Capability}
\label{table:Elo_Average_01}
\end{center}
\end{table}

\begin{table} 
\begin{center}
\begin{tabular} {| c c | c |  }
\hline
 Agent X & Agent Y & $P[Y \succ X]$ \\
\hline
 \color{myDarkRed}1370 &  \color{blue}1400   & 0.5431 \\
 \color{blue}1200 &  \color{myDarkRed}1100   & 0.3599 \\
\hline
\end{tabular}
\caption{Average Ratings of with Two Mixed Capabilities}
\label{table:Elo_Average_02}
\end{center}
\end{table}

Because they model real combat, where different sides
face different conditions, this example is particularly relevant for paper strategy games.
Suppose we have a game in which Red and Blue start from different
sides of the map and thus face different terrains, as well as potentially have different
initial forces. One might expect that a single agent might exhibit
different capabilities in various situations, based on their experience and skill. Suppose agent X can
play Red and Blue equally well, e.g. with Elo rating 1200. Further suppose that,
because of some quirk of software,  agent Y can play Red with level 1100 but can play Blue with level 1400.
Perhaps the geometric terrain analysis capabilities of Y just work better on the kind of terrain
prevalent on Blue's side of the map.

What overall Elo rating should Y have? According to equation \eqref{eq:Elo_Prob_Formula}, the difference
in rating is determined solely by Y's probability of defeating X. Table  \ref{table:Elo_Average_00} shows
that if X plays Red and Y plays Blue, then
the $1200\!:\!1400$ means Y as Blue has $p_B =0.7597$ probability of winning. If the colors are reversed,
then Y as Red  has $p_R=0.3599$ of winning. If roles are assigned randomly,
then both combinations are equally likely and  Y's overall probability of winning is
just the average probability $p=(p_R + p_B)/2 = 0.5598$ of winning, which corresponds to an Elo advantage of
about $41.8$ points, so Y should have rating about $1242$.  Table  \ref{table:Elo_Average_01} shows
that if Y's ability to play Blue were increased to 2500,
then the overall probability of winning would be $p = 0.6797 = (0.3599 + 0.9994)/2$, which
is equivalent to an overall rating of about $1331$. It is worth noting that the arithmetic, geometric, and harmonic
averages of 1100 and 2500 are 1800, 1658, and 1528 respectively, so even though the "elo-average" is clearly 
defined, it is also clearly  something quite different from  those three.

A more realistic situation is that the abilities of both X and Y vary according to which color they play, as in table
 \ref{table:Elo_Average_02}. In this case, Y's overall probability of winning is $0.4515$, so equation \eqref{eq:Elo_Prob_Formula}
specifies that X has an Elo advantage of about $33.8$ points. Specific levels for X and Y would have to be assigned
by taking into account their performance against a wider variety of opponents in a tournament.

\subsubsection{Averages for Initial Ratings}

Another application  might be to assess the average
rating of a population of opponents, given how one player fared.
This is the converse of the usual application. Usually, we want to estimate the rating
of an individual, assuming the collection of opponent ratings is known and fixed.
In this case, we want to estimate the average rating of a collection of opponent ratings,
assuming that the rating of an individual is known and fixed. As AI agents do not learn or change
after they are trained, but new unrated agents are generated, this second case is useful for
machine learning but non-existent for human play.

Suppose we have a player $X$ (human or software) with a rating of $ X = 1320$, who won an average of
$p_1 = 0.675$  of their games. We need to determine the average opponent rating, $A$ to match that probability.
We use the formula of equation  \eqref{eq:Elo_Prob_Formula} 
in each
term of equation \eqref{eq:Generalized_Averages} to define a new generalized average. 
Equation   \eqref{eq:Elo_Prob_Formula} defines each individual term of the sum.
In this case, the $R_i$ terms are being averaged:

\begin{equation}
f(R_j, X) = \frac{e^{\beta X}}{e^{\beta X} + e^{\beta R_j}} 
\end{equation}

We substitute this into equation \eqref{eq:Generalized_Averages},
noting that the overall probability of victory is the arithmetic 
average of the probability against each particular opponent:

\begin{eqnarray}
 \sum_{j=1}^N  \frac{e^{\beta X}}{e^{\beta X} + e^{\beta R_j}} &=&  \sum_{j=1}^N  \frac{e^{\beta X}}{e^{\beta X} + e^{\beta A}}  \nonumber \\
N p_1 &=&  N  \frac{e^{\beta X}}{e^{\beta X} + e^{\beta A}}  \nonumber \\
   p_1 &=&     \frac{e^{\beta X}}{e^{\beta X} + e^{\beta A}}  \nonumber \\
&=&     \frac{1}{1 + e^{\beta (A-X)}}  \nonumber \\
\frac{1-p_1}{p_1}&=&     e^{\beta (A-X)}     \nonumber  \\
A - X &=&  \frac{1}{\beta} \ln\left( \frac{1-p_1}{p_1} \right)
\label{eq:Advantage_from_Log_Odds}
\end{eqnarray}

Finally,
\begin{equation}
A_1 = X + \frac{1}{\beta} \ln\left( \frac{1-p_1}{p_1} \right)
\label{eq:Population_Mean_Rank}
\end{equation}

Substituting the provided numbers, we see that the right hand term is $-127$ points,
so the average rating of the opponents is $A_1 = 1193$. Again, we note that the
probability of winning or losing only determines the difference in Elo ratings.
The Elo advantage is strictly proportional to the log-odds of victory.

The primary use of this equation is to determine the Elo rating of a new population.
For example, a genetic algorithm could
use the probability of success as the evaluation metric for hundreds of agents,
using some agents of known capability as the training opponents.
After convergence, we would only want the Elo rating of a dozen or so of 
the best, and we would have their recorded   probabilities  to start the SC-Elo
calculation.
If the new probability fell to $p_2 = 0.423$,
then we could put the new value into equation \eqref{eq:Population_Mean_Rank}
and immediately see that the new population should have average Elo rating
of $A_2 = 1374$ with individual agent ratings spread around this value.

A useful application of equation  \eqref{eq:Population_Mean_Rank} is to estimate
the improvement in capability between two populations (e.g. two different eras of a genetic algorithm)
that have some overlap. Suppose that there are $M+N$ in both populations, where
$N$ agents appear in both populations. We denote $p_1$ as the average win rate of those
$N$ agents when playing against the $M$ agents only in the first populations.
Similarly, $p_2$ is the average win rate of those same $N$ agents when playing against the $M$ agents only
in the second population. Then we can take the difference of equation  \eqref{eq:Population_Mean_Rank},
so that $X$ cancels,
to get the average Elo improvement of the second population over the first:

Finally,
\begin{eqnarray}
A_1 - X &=&   \frac{1}{\beta} \ln\left( \frac{1-p_1}{p_1} \right)  \nonumber  \\
A_2 - X &=&   \frac{1}{\beta} \ln\left( \frac{1-p_2}{p_2} \right)  \nonumber  \\
A_2 - A_1 &=&  \frac{1}{\beta}  \left[
 \ln\left( \frac{1-p_2}{p_2} \right)
 -
 \ln\left( \frac{1-p_1}{p_1} \right)
 \right]
\label{eq:Population_Rating_Improvement}
\end{eqnarray}

The simplest case is when $M = 1 = N$. If agent X wins against Y 75\% of the time, then X is 200 Elo points
over Y. If X wins against Z only 25\% of the time, then X is 200 Elo points below Z. Therefore, Z is 400
Elo points higher than Y.

 In this way, we can accurately determine the initial rating estimates 
for the new agents, which is necessary to start the SC-Elo procedure and get individual ratings
for each new agent. To simply use a default value like $1000$ would be off by hundreds
of points, and thus incorrectly depress the estimated ratings.
The problem would get worse with each succeeding generation of improved AI agents.

\subsection{Precision}
\label{sec:Precision}

When working with Elo ratings, how precise should we be? In the iterative numerical algorithms described in this paper,
the question arises of when to stop refining the estimated ratings. With 64-bit arithmetic, 
dozens of decimals can be calculated, but they probably are not significant.  In training artificial intelligence,
we must determine what is a significant performance improvement, and decide how to detect it.

As with any statistical problem, we can only get probabilistic answers. We will use a binomial model. In $N$ games
where agent X has (unknown) probability $p$ of defeating Y, the expected number of victories will
be $pN$ with variance $p(1-p)N$. Hence, the expected fraction of victories will
be $p$ with standard deviation $\sqrt{p(1-p)/N}$. We will say that $N$ trials  is enough
if the expected difference between the mean when $p_1=0.5$ and $p_2 > 0.5$ is 
at least a given multiple of the standard deviation of that difference.\footnote{We make
use of the fact that the variance of the difference between two random variables is
the sum of their variances.}  For the fairly 
large $N$ in table \ref{table:Sample_Size}, the errors in the observed frequencies are approximately
normal, so the probability of  being correct with $2\sigma$ is about $0.977$,
with $3\sigma/2$ about $0.933$, and with $\sigma$ about $0.841$.

\begin{table} 
\begin{center}
\begin{tabular} {| c c  | c c c |  }
\hline
  Advantage & $p_2$ &  $\sigma$ &  $3\sigma/2$ &  $2\sigma$ \\
\hline
1        & 0.5014 &   241,423 &  543,202 &  965,693 \\
5        & 0.5072 &       9,657 &    21,729 &    38,629 \\
10      & 0.5144 &       2,415 &      5,433 &      9,658 \\
25      & 0.5359 &          387 &         870 &      1,546 \\
34.9   & 0.5500 &          199 &         448 &         796 \\
50      & 0.5715 &            97 &         218 &         388 \\
100    & 0.6401 &           25  &          55  &           98 \\
150    & 0.7034 &           11  &          25  &           44 \\
200    & 0.7596 &             6  &          14  &           26 \\
\hline
\end{tabular}
\caption{Required Sample Sizes}
\label{table:Sample_Size}
\end{center}
\end{table}

\begin{figure}[ht]
\begin{centering}
\fbox{\includegraphics[scale=0.75]{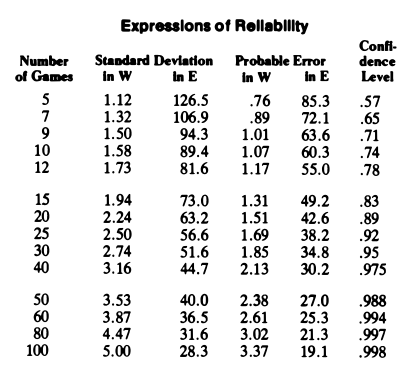}}
\par\end{centering}
\protect\caption{Elo's Expressions of Reliability}
\label{fig:elo-reliability} 
\end{figure}

Clearly, for a pair of humans to play a hundred games in a row is infeasible, so one
might easily consider Elo differences of 50 points or less to have very little practical significance
for human ratings. As of early 2021, the two highest Elo ratings ever achieved were 2882
for Magnus Carlsen and 2851 for Garry Kasparov. The 31 Elo point difference corresponds
to  only a 54.45\% chance of the strong player winning one game. To get a one standard
deviation difference between that and an even game would take about $250$ games in a row.
While the format has changed frequently, the Chess world championship has recently decided
by an 8-player double round robin to pick the challenger, then a 14-game match between the
challenger and reigning champion.

However, two software agents could easily play a few thousand games, so differences
of around 10 points are still feasible to detect, though probably only differences
of 20 or more should be considered meaningful.
Considering the row with Elo-advantage of $34.9$ points, it is not surprising
that the AlphaGo system accepted a newly trained agent as a meaningful improvement
if it achieved at least 55\% wins in 400 games, i.e. $1.4$ standard deviations 
difference, or at least 35 Elo points with 92\% probability.

Elo presented his system  in \cite{RatingChessPlayers}. In Figure
\ref{fig:elo-reliability}
we reproduce a table from section 2, paragraph 55, which presents
the uncertainty in Elo ratings for various sizes of tournaments. While 
the exact formula Elo used is unclear, his point was that it would
take a  large number of games between the same two players
to reliably determine an Elo advantage of a few dozen points.
He then emphasized that a 15-round tournament
has a probable error of 49.3 points, which sets a bound on
how precise a result can be expected, simply because tournaments are
small samples from a binomial process.

\subsection{Group Total of Ratings}
\label{sec:Total_of_Ranks}

When a group of players are all rated using the same system, subtle group dynamics emerge
which present fundamental problems. For groups of people, the issues appear
as rating inflation or deflation over time, fairness to ongoing players when others
join or leave, and the possibility of perverse incentives in choosing with whom to play.

For groups of software agents, the most basic problem is tracking a rising level of skilled
play when each new generation of improved agents have similar probabilities of winning.
For example, in the AlphaGo system, the early AI agents played almost randomly, and 
each had roughly 50\% win rate against other semi-random agents of the same era.
Later in training, the AI agents might have the strength of a human grand master,
but still have roughly 50\% win rate against other agents of the same era.
When the only measure of performance is the win rate against a population, the
win rates cannot be meaningfully compared when the population changes over time.

The root of the problem goes back to the fundamental rating formula.
In both the classic and self-consistent Elo systems, for rating players in a tournament,
each player $i$ gets updated ratings according to the usual formula:

\begin{equation}
R'_i = R_i + K_i (N_i - E_i)
\end{equation}

For the following analysis, it turns out that the details of how $E$ is estimated (specifically,
classic Elo or self-consistent Elo) do not matter.
The $N_i$ term is the sum over opponents of $i$'s actual wins over each opponent: 

\begin{equation}
N_i = \sum_j N_{ij}
\end{equation}

Of course, for all $i,j$ pairs,  $N_{ij} + N_{ji} = 1$

The $E_i$ term is the sum over opponents of $i$'s estimated probability of wins over opponents,
whether the initial are used in classic Elo or final ratings are used in SC-Elo:

\begin{equation}
E_i = \sum_j E_{ij}
\end{equation}

Regardless of how they are estimated, for all $i,j$ pairs,  $E_{ij} + E_{ji} = 1$.

If we denote the discrepancy between performance and expectation as $d_{ij} = N_{ij} - E_{ij}$, then it is clear that
for any $i,j$ pair that played each other in the tournament, $d_{ij} + d_{ji} = 0$. Hence, the sum over
all $i, j$  pairs that did play each other is also zero. We can simplify the summation indices without changing the
sum  by adding zeros for all those
pairs that did not play each other. For that purpose,  we adopt
the convention that $d_{ij} =d_{ji} = 0$ when $i$ and $j$ did not play each other\footnote{When no play is observed,
no discrepancy in outcome can be observed.}. We
can rewrite the rating-update equation as a sum over all players:

\begin{equation}
R'_i = R_i + K_i  \  \sum_j d_{ij}
\end{equation}

If all the players had $K_i = K$, then we could demonstrate that the sum of the final ratings
was the same as the sum of the initial ratings:

\begin{eqnarray}
\sum_i R'_i  &=& \sum_i  R_i + K \  \sum_i  \sum_j d_{ij}  \nonumber  \\
                      &=& \sum_i  R_i + K \   \sum_{i<j} \left( d_{ij} + d_{ji} \right)   \nonumber \\
                      &=& \sum_i  R_i 
\end{eqnarray}

Because the $d_{ij}$ terms spread out of the whole group of players, it is perfectly possible 
in a large group for a few 
excellent players to increase their ratings significantly - but the constant-sum effect
guarantees that all the rest will have exactly compensating small declines in their ratings.
In short, the constant-sum condition on each individual pairwise contest
extends to the whole group because addition is associative.

The practical implication of this result for strategy games is that it will be difficult to incrementally update
the Elo ratings in a small group (either human players of strategy games or AI agents being trained). If raising
the rating of one player means an exactly corresponding decrease spread over
the ratings of just a few others, it is impossible
to show a series of agents increasing in Elo rating over time: that would require the sum of their ratings
to increase over time, which we have just shown is impossible unless the formula is modified somehow.

With a large, diverse group  of players who each play a wide variety of opponents, the constant-sum effect
is  concealed by the large direct effects of  statistical noise of games, changing levels of skill as people improve their
game, players joining or leaving the group, and so on. 
Nevertheless, the effect would still show up at the group level over time, and there are ongoing debates
in the chess community about exactly what might or might not be happening, and what (if anything) 
should be done.

Most techniques to deal with this problem involve carefully changing the $K$ values
so that the sum of the ratings not only does change, but does so in a way that
is perceived as fair by human players and accurately measuring the changing group performance
of software agents.

As of February 2021, the FIDE rules  have several pages of detailed and complex calculations
for computing different effective $K$ values for each player, but the general intent is to make $K$
larger for lower-rated players whose skills are less well known and more likely to be improving rapidly
and smaller for higher-rated players whose skills are well known and less likely to be changing.

If the $K$ were set purely by the ratings, then play between higher and lower rated players would
tend to consistently raise the sum. The following analysis indicates that this is due to a
second-order effect whereby giving low $K$ to high-rated players produces asymmetric
errors in the estimated win-probabilities. While that sounds like a technical error, it is actually
necessary to assure fair treatment of experienced players when facing low-rated opponents.
It also helps stabilize the whole system, thus making the ``average Elo rating'' a more useful
point of comparison, by making the ratings of the most skilled players be the most stable ratings.

For just two players, we will call the condition that $K_B \ge K_A$ whenever  $R_A > R_B$
the ``KR-condition''. For a whole population, the KR-condition is that there be a 
negative correlation between $K$ and $R$, but not necessarily an absolute guarantee
for every pair.
Suppose we have two players, A and B,  where the KR-condition holds.
Given those ratings, the Elo formula predicts a probability
 that A wins which we call $Q$. The actual observed frequency of a victory by A is $f$.

In the case of victory by A, we have the following expression for the change in A's rating:

\begin{eqnarray}
R'_A  &=& R_A + K_A(1-Q) \nonumber  \\
\Delta A_V &=&K_A (1-Q)
\end{eqnarray}

Similarly for B's rating:
\begin{eqnarray}
R'_B  &=& R_B + K_B(0-(1-Q)) \nonumber  \\
\Delta B_V &=& -K_B (1-Q)
\end{eqnarray}

Adding, we get the change in the sum when A wins:
\begin{equation}
\Delta S_V = -(K_B - K_A) (1-Q)
\end{equation}

Regardless of  what system  is used to produce the estimate $Q$, the total of the ratings will fall when A wins if $K_B > K_A$.
Similar expressions apply in the case of a loss by A:

\begin{eqnarray}
R'_A  &=& R_A + K_A(0-Q) \nonumber  \\
\Delta A_L &=& -K_A Q
\end{eqnarray}

\begin{eqnarray}
R'_B  &=& R_B + K_B(1-(1-Q)) \nonumber  \\
\Delta B_L &=& +K_B Q
\end{eqnarray}

\begin{equation}
\Delta S_L = +(K_B - K_A) Q
\end{equation}

The total of the ratings will rise when A loses if $K_B > K_A$.

\begin{eqnarray}
E[\Delta A] &=& f \Delta A_V + (1-f) \Delta A_L  \nonumber  \\
                &=& K_A \left[  f(1-Q) - (1-f)Q  \right]
\end{eqnarray}

Similarly,
\begin{equation}
E[\Delta B] =  K_B \left[  f(1-Q) - (1-f)Q  \right] 
\end{equation}

Given that A wins with frequency $f$, the expected changes are the weighed sums. As one would expect,
when the ratings accurately 
reflect current strengths, $Q = f$,  both A and B  will (on average) have no change in their ratings. In particular,
we can set $K$ within broad limits to achieve other goals, so long at the KR-condition holds.
The expected change in the sum is therefore the following:

\begin{equation}
E[\Delta S]   = \left[ K_B -  K_A  \right] \ \left[  (1-f)Q -  f(1-Q)   \right]
\label{eq:Expected_sum_change}
\end{equation}

The first term in the right hand side of equation  \eqref{eq:Expected_sum_change}
is never negative because  $K_B \ge K_A$.
The second term is positive when $Q > f$. Thus, the average rating will rise
over time if the probability of strong players defeating weaker ones is overestimated
more often than not.

Exactly as before,
the positive expectation  on each individual pairwise contest
extends to the whole group because addition is associative.
This would tend to produce an overall slow rise in the sum of the ratings, except for the
contrary effect of entry and exit. New players enter the system with low ratings while
senior players exit the system with higher ratings, which tends to pull down the average.\footnote{
More definitive analysis to explain the changes in 
average chess ratings would have to rely on empirical data to settle several outstanding
issues. Merely comparing the average of recorded ratings would have
to deal with confounding issues. For example, with increasing online play over the
past few decades, it is entirely possible that a different population of players - strong or weak -
can now  play who were previously excluded, which would make comparisons
between years problematic because they might be based on quite different populations.}

One way to ensure that the KR-condition, and hence $Q > f$, holds
  in practice is the use of provisional ratings. For some online game forums,
there is a certain period (e.g. twenty games) in which the new player B's rating is rapidly adjusted
(large $K_B$)
but the scores of their  opponent is never altered (zero $K_A$)\footnote{When
two provisional players have a game, the result affects neither rating.}. Without the provisional ratings,
experienced players who are new to the forum would come in with the lower rating of a true
novice, which would then rapidly rise at the expense of other players. The constant-sum problem
would show up as unfair lowering of the ratings of good players just because a new one joined.

With provisional ratings, talented players who are new to the forum can rapidly receive a realistic initial rating
and experienced players run no risk by welcoming newcomers.
For example, suppose a new player had the skills of an Elo-1500 player but was assigned 
the standard novice rating 1000. An experienced Elo-1700 player 
would
be rapidly pulled down, even though they were actually better than the new player.
Because the sum of the ratings would stay at 2700, the ratings would equilibrate at 1250 and 1450 
respectively, an outcome unfair to both.
The two players with artificially low ratings would then play others, and pull them down
as well until those two had risen up to realistic ratings - but only by pulling down all the others slightly.
Temporarily using $K_A = 0$ eliminates this unfair outcome.  In our example, the initial situation of $Q=0.95$ but
$f = 0.76$ would cause the new player's rating to rise until $Q=0.76=f$, without penalizing the better player.
The new ratings would equilibrate at 1500 and 1700 because the sum was allowed to rise from 2700 to 3200.
During the equilibration period, $Q > f$ would hold
and the sum would increase. This example shows that $Q > f$ is not a permanent problem with
the rating system but simply an accurate description of the transient equilibration period.

To avoid the constant-sum problem  for non-novices,
the same systematic variation of $K$, and hence $Q > f$, must apply
to the non-novice players as well. 
 For a very highly-rated A facing a highly rated B,
the same process would happen again (but at a higher level) because A's rating
would adjust downward
more slowly than B's adjusted upward. This preceding analysis
 indicates that the  KR-condition  is important for fairness at the low, middle, and high
ratings - which is exactly  the FIDE system  mentioned 
at the start of section \ref{sec:Adjusting_Variance}. The stated rationale
is that the quality of play is better known for players with long records, but it also
turns out that decreasing $K$ with $R$, i.e. the KR-condition, is necessary to allow
excellent players to reach a rating equal to their 
actual capability without pushing good players below theirs.

The fact that the KR-condition can escape the constant-sum problem 
for even small  groups leads to the following suggestion for AI in strategy games.
When training agents with  population-based methods (e.g. genetic algorithms), we could rate them in batches as follows.
First, a set of hand-written scripts are produced and assigned a fixed, permanent rating of Elo 1000 with $\sigma = 0$
so no adjustment occurs. 
In the first era, we might generate $N=50$ random agents and train them against the scripts. We can then rate
the new agents and scripts together in a batch, starting with  $R=1000$ and $\sigma=1000$ for the new agents while
keeping $\sigma = 0$ for the scripts. The effect is that the new agents have provisional ratings that change rapidly
without affecting the rating of non-provisional players (the scripts).

We might then
use the $K=10$ best agents as a fixed set of opponents against which to train the next $N$ agents. We could then
rate the new agents and old agents together in a batch, keeping the $R$ of the old agents via $\sigma=0$. 
The second generation of agents are given provisional ratings that change rapidly, without affecting the
ratings of the non-provisional players (previous generation).
In this way,
each set of agents has its Elo ratings anchored by the rating of earlier agents, all the way back to the scripts.

As equation   \eqref{eq:Expected_sum_change} indicates, any system in which $A > B$ implies $K_A < K_B$
would lead to gradual increase in ratings over time, because of the second order effect that $Q > f$. 
Many such systems could be designed; the above are just a few suggestions.

The analysis of this section applies only to a series of incremental updates such as that experienced
by a person playing a series of tournaments over their chess career. When all the ranks are adjusted
simultaneously in one batch job, as discussed in section \ref{sec:Batch_Adjust}, there is no
way to determine - before the ratings are calculated - which agents should have large $K$ or small.
All that is necessary (or possible)
is that all agents have an uninformative prior, i.e. large $\sigma$.

\clearpage
\section{Simultaneous Adjustment of Ratings}
\label{sec:Batch_Adjust}

As mentioned in section \ref{sec:Adjusting_Elo_Ranks}, adjusting a single rating under the approximation  that all
other do not change is convenient, but leads
to problems in some circumstances (e.g. 382 point gap when 191 is correct). 

Humans always mix learning and performance. Young players develop their skills, old players lose their edge,
everyone gains experience with every game, and skills get rusty when not practiced. 
Thus, constant incremental update of ratings after every tournament
(more generally, after every $N \ge 1$ games) makes sense for tracking their ever-changing abilities.
However, many software agents are completely static: once they have been trained, their abilities
do not change. Of course, a new agent might be developed by altering the neural network weights of an agent,
but that would be the creation of a second agent: the performance of the initial set of weights would be unchanged.

In this sense, AlphaGo is best viewed not as a single agent that got better over time but as a
system for producing a sequence of ever-improved agents. While Lee Sedol played a particular agent,
using a particular set of weights in a neural network, the graphs of improving performance over time have a different
agent's performance at each moment - even though all were produced by the AlphaGo system.

This presents an argument for simultaneous updates over incremental updates. Suppose that agents A and B both 
have ratings 1100. Suppose that A defeats B in 4 out of 5 games, which indicates roughly 250 Elo point advantage.
Because of the anchoring effect of prior uncertainty, they would adjust by considerably less.
Suppose they both shift 50 points and are assigned new ratings of (e.g.) 1150 and 1050.
Over the next few hundred games, B loses to Elo-1000 agents 76\% of the time, so that B's rating 
gets adjusted to 800 over time. But because B's true ability was the same all along, this means that A actually
defeated an opponent with a rating of just 800. However, A still retains the 1150 rating, as
if it had defeated an opponent of rating 1100. Even though
the 4 out of 5 result was clearly based on a random fluctuation, there is no way to go back and revise A's rating based
on later information. While retroactive adjustment would be manifestly unfair to humans whose performance
changes constantly, it seems only logical for software agents with static performance.

Simultaneous batch update fixes this problem. As all the games of A, B, and all others are updated at once,
the later performance of B is used to set the rating against which A's early play is compared. Indeed,
the order of games is not used and does not matter. This is exactly the same situation as when using
equation \eqref{eq:Classic_Elo_Update} to update ratings after a FIDE tournament: the order of games inside
the tournament is not used and does not matter.

Analytically, the solution is simply to repeat the entire derivation of the maximum posterior likelihood, except that the joint
probability of all $R_i$ is maximized. The same formula for the PML estimate of each $R_i$ is obtained,
except that the expected win rate is based on all the new ratings:

\begin{equation}
R_i = \mu_i + \beta \sigma_i^2 \left( A_i - E_i (R_1 ,  \ldots , R_M) \right), \  \   1 \le i \le M
\label{eq:Vector_SC_Elo}
\end{equation}

Because the posterior probability of $R_i$ is influenced by the prior, the formula works even if a particular sub-group has no
games linking them to another sub-group. Indeed, if some $R_k$ is completely isolated and has no games with any other player,
then the $A$ and $E$ terms are summed over zero opponents, and we get $R_k = \mu_k$ for that player: 
no new information quite reasonably leads to no change and does not cause the algorithm to 
produce nonsense or even crash with an error.

\begin{figure}[ht]
\begin{centering}
\includegraphics[scale=0.5]{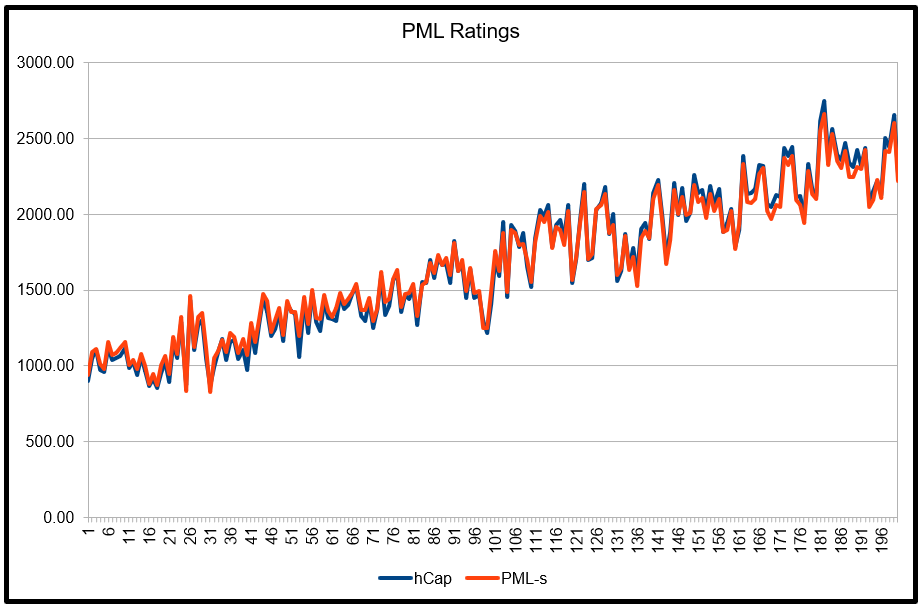} %% this is the name of the image file
\par\end{centering}
\protect\caption{True capability and PML-fitted Elo ratings}
\label{fig:PML-fitted-ranks}  %% label by which it is referenced 
\end{figure}

As before, numerical stability is a problem, so the slope must be taken into account.
The end result is much the same as equation \eqref{eq:Consistent_Elo_Update_Algorithm}, except that the new 
$E_{i,t}$ and $s_{i,t}$ calculations use
not only the new $R_{i,t}$ but all the other new $R_{j,t}$ values as well.
Because different players should usually have different $K$ values, such as provisional players
with $K=0$, the algorithm allows different $K$ values for each agent.
Equation \eqref{eq:SC_Elo_Height} for $E_i(r)$ and
the same  \eqref{eq:SC_Elo_Slope} for $s_{i,t}$ use the ratings of all the opponents
already, so the only change in software is to remember to update them at each
iteration of the batch fitting algorithm.

\begin{eqnarray}
K_i &=& \beta \sigma_i^2 \\
 s_{i,t} &= & \frac{\partial E_i}{\partial r_i}\Bigr|_{\substack{R_{i,t}}} \\
\hat{R}_{i,t+1}  & = & \frac
 {R_{i,0}  \ +  \ K_i \left[ A_i - E(R_{i,t}) + s_{i,t} R_{i,t}  \right] }
 {1  \ + \   s_{i,t} K_i }  \\
R_{i,t+1} &=& \frac{ \hat{R}_{i,t+1} + R_{i,t}} {2}
\label{eq:Consistent_Elo_Batch_Update_Algorithm}
\end{eqnarray}

The variable batch size $M$ solves several problems. First, for $M=1$ we would just update one agent's rating, holding
all the others constant. For $M=2$, we can adjust two opponents so as to avoid creating a double-sized gap between them.
The C++ code for fitting was tested with $M=1$ and $M=2$ in cases which could easily be verified by hand-calculation.

To test that no unexpected problems arose on larger cases, we generated a series of 
$M=200$ simulated agents in 10 eras of 20 new agents each with slowly increasing capability each era.
Note that these "agents" had no real capability
to play, just assigned numbers to designate capability: the purpose was to test fitting, regardless of how the
data was generated.

In each era, the five agents which had
the highest win percentage in the previous era (if any) were carried over so that all eras
after the first had 25 agents playing each other. Each new agent played 100 games as Red against
the carry overs and 100 as Blue. Thus, each of 20 agents played 200 games, so each era involved
4000 simulated-games.

On average, $20/25$ of the agents in an
era did not make it into the next era. While it was typical for five of those agents
to be carried over from the previous era, some highly capable agents got carried over twice.

To ensure that the fitting procedure did not implicitly depend on the agents having
exactly the form in equation  \eqref{eq:Elo_Prob_Formula}, we designed the
simulated agents so that it was difficult to clearly define exactly what the "true capability" of
each agent might be.
Similar to section \ref{sec:Mixed_Capabilities}, each agent was assigned different capabilities
when playing Red and when playing Blue. The average capability was raised by 150 Elo points
each era to simulate the effect of training new AI agents. Each agent was assigned  the level
for that era, plus or minus 200 points. That base capability was perturbed by plus or minus 200
points to get the Red capability, and similarly for Blue. Thus it was possible (but unlikely) for an agent to have both
capabilities 400 points above the average for that era or up to 400 points below the average. Similarly,
it was possible for the Red and Blue capabilities of an agent to differ by as much as 400 points.
As illustrated in section  \ref{sec:Mixed_Capabilities}, there is no obvious formula for combining
Red and Blue capabilities of an agent to get an overall capability that precisely predicts the
overall win probability against a fully-specified opponent.

To play one agent against another, they were randomly assigned the Red and Blue roles, then
their respective Red or Blue capabilities were used in equation \eqref{eq:Elo_Prob_Formula} to
stochastically determine a winner for that game.
This stochastically generated a series of 40,000 game-outcomes.
Note that the random draws according to equation \eqref{eq:Elo_Prob_Formula} will inevitably introduce
statistical noise, preventing a perfect fit.

Fitting 200 ratings
from 40,000 games 
all in one batch update produced  Figure \ref{fig:PML-fitted-ranks}. The blue line shows the capability
for each of the 200 simulated agents in the order they were created. We simply combined the Red and
Blue capabilities of each agent using the harmonic average, for reasons discussed in section 
 \ref{sec:Mixed_Capabilities}. 

The red line shows the ratings fitted by equation \eqref{eq:Consistent_Elo_Batch_Update_Algorithm} 
 based on all 40,000 games; the correlation is over 99.7\%.

To start the fitting algorithm, some prior information about agent ratings must be used. Because we knew
little except  that there would be a wide variety of capabilities, an uninformative flat prior distribution was used:
every agent was modeled has $\mu = 1000$ with $\sigma = 1000$. In future work,
the methods of section \ref{sec:Posterior_Variance_Estimation} could be applied to estimate the posterior uncertainty.

In light of section \ref{sec:Precision}, the termination criterion on the fitting was that  every
estimated rating fall within the tolerance  $| \hat{R}_{i,t+1} - R_{i,t} | < 0.05 $. This provides one decimal accuracy in the Elo ratings,
which is both far more precise than is needed  for Elo ratings and far less precise than can be achieved
with 64 bit double precision numbers. While convergence only takes a few iterations when $M=2$, several
thousand iterations were required when $M=200$.  One of the reasons for the increase is the
need to propagate influence from the very earliest ratings to the later. Because the nearly-uninformative
was centered on 1000 points, the fitting algorithm tended to make the middle ratings near 1000, the early ones
considerably less, and the later ones considerably more. For some runs, this caused some of the early ratings
to hit the absolute minimum of 100 Elo points. To make the difference in ratings correspond to the 
victory probabilities, it was therefore necessary to raise all the ratings. Propagating this influence
from the early ratings 
across hundreds of rating to the later ones required more iterations.

As has been frequently mentioned, only the difference in Elo ratings has meaning in terms of probabilities. Absolute
values of Elo ratings are meaningless (except when compared to a large population with a stable average, like
the chess community). A set of ratings can be shifted up or down, but may not be rescaled. Adding 1000 to all
Elo ratings would not change the Elo-advantages, but scaling them all up by 50\% would 
raise the Elo-advantages. A 100-point advantage would become 150, while 200 would become 300: not only would
the probabilities change, but they would change differently for each case. To make it visually obvious
that the Elo-differences in red fitted ratings closely matched the Elo-differences in the blue capabilities, we shifted
the fitted ratings (without any rescaling) to make the means coincide.

When training agents with  population-based methods (e.g. genetic algorithms), we could rate
 them in batches as follows.
In the first era, we might generate $N=50$ random agents and train them against hand-written scripts. We can then rate
the new agents and scripts together in a batch, setting the $R=1000$ and $\sigma=0$ for the scripts. We might then
use the $K=10$ best agents as a fixed set of opponents against which to train the next $N$ agents. We could then
rate the new agents and old agents together in a batch, keeping the $R$ of the old agents with $\sigma=0$. In this way,
each set of agents has its Elo ratings anchored by the rating of earlier agents, all the way back to the scripts.

\clearpage
\section{Basic Assumption of Elo Ratings}
\label{sec:Elo_Assumptions}

While equation \eqref{eq:Elo_Prob_Formula} is usually cited as the foundation of the rating system,
Elo's development of the equations did not start there.
Elo presented basis of his system in \cite{RatingChessPlayers},  section 8, starting at paragraph 8.33,
which we quote here with some reformatting of his equation:

\begin{quote}
Assume three chessplayers, x, who has a certain chance - odds in his favor - of scoring over y,
who in turn has some other odds of scoring over the third player z. The odds of x to score over
z are
\end{quote}

\begin{equation}
\left( \frac{P_{xy}}{P_{yx}} \right)
\left( \frac{P_{yz}}{P_{zy}} \right)
=
\frac{P_{xz}}{P_{zx}} 
\label{eq:Player_Odds_Chain_Rule}
\end{equation}

Taking logarithms, Elo introduces the quantity we called the ``Elo advantage'' in equation 
\eqref{eq:Advantage_from_Log_Odds}, where his $C$ is our $1/\beta$.

\begin{equation}
C \ln \left( \frac{P_{xy}}{P_{yx}} \right) = D_{xy}
\end{equation}

He notes that equation \eqref{eq:Player_Odds_Chain_Rule} obviously implies that differences
in rank are additive:

\begin{equation}
D_{xy} + D_{yz} = D_{xz}
\label{eq:Additive_Advantages_Rule}
\end{equation}

He then  builds the standard Elo scale of ratings and equation \eqref{eq:Elo_Prob_Formula}, all based on 
assumption  \eqref{eq:Player_Odds_Chain_Rule}. In the rest of this paper,
we will use $A_{xy}$ for the observed Elo-advantage of $X$ over $Y$, not $D_{xy}$.

\subsection{Non-Multiplicative Odds}
\label{sec:Nonmultiplicative_Odds}

However, it should be noted that equation \eqref{eq:Player_Odds_Chain_Rule} is a very strong assumption,
and it is easy to generate plausible examples which violate it: the odds of victory are not always multiplicative
so the differences in ratings are not always additive
as equation \eqref{eq:Additive_Advantages_Rule} requires. Elo himself, and many others since, have
cautioned that while Elo ratings can be calculated from the unbroken records of chess games going back hundreds 
of years, comparison of ratings from different generations are largely meaningless. The reason is that
chess is a complex game, there are different styles that utilize different areas of capability, and the styles of
play have changed greatly over time. While it is easy to compare players from the early 1800's to each other
and to compare players from the early 2000's to each other, it is not at all clear how either one would fair against the other.

While many articles have been written in debate over this topic by many authors,
we present a simple algebraic example of how this can happen, based on the Condorcet Voting Paradox
\cite{Condorcet}. This is not really 
a paradox at all, but just a surprising example of how majority rule will typically produce circular preferences
in groups when more than two  voters have preferences over more than two candidates\footnote{A similar situation
can happen when trying to attribute ``combat worth'' to weapon systems that exhibit the
circular pattern of rock-paper-scissors dominance. A real-world example can be found in the early 1960's
when some American strategists assumed that their (expected) ability to defeat the Russian superpower
would automatically translate into the ability to easily defeat the Vietnamese regional power.
However, it turned out that the capabilities for  fighting high-intensity conflict against
the Soviet Union were  qualitatively different from  the capabilities required
to wage guerrilla war in Vietnam.}. The simplest way to
produce a counter-example to equation  \eqref{eq:Player_Odds_Chain_Rule} is to explicitly
include three or more different areas of capability, like players from different eras or the voting paradox.

\begin{table} 
\begin{center}
\begin{tabular} {| r | r r  r |  }
\hline
 & Tesuji & Joseki & Ko \\
\hline
Alice   & 300 & 600   & 900 \\
Bob    & 900 & 300   & 600  \\
Carol  & 600 & 900   & 300  \\
\hline
\end{tabular}
\caption{Circular Ratings}
\label{table:Circular_Ratings}
\end{center}
\end{table}

In table \ref{table:Circular_Ratings},
we have three Go players. Each has some level of capability in three areas, nominally
tesuji, joseki, and ko. If a player out-performs their opponent in at least two out of three
areas, they win that game. The numbers in the matrix show the 
capability for each player in each area, stated as (ironically) Elo ratings in each area.
Thus, Alice has a 0.849 chance of out-performing Bob in joseki, a 0.840 chance  in ko,
but only a 0.031 chance of out-performing him in tesuji. This works out to a 0.729
probability of out-performing him in two or three areas. By equation \eqref{eq:Advantage_from_Log_Odds},
this probability of victory is an overall Elo advantage of $A_{AB} = +172$ points\footnote{More precisely, 
$A_{AB} = 171.635$, but
we have rounded; the numbers in this and other examples  might  not add up exactly.}.
Similarly, we can see that Bob is likely to beat Carol, because he is
better at both tesuji and ko: $A_{BC} = +172$.

According to equation \eqref{eq:Additive_Advantages_Rule}, Alice should have a $+343$ Elo advantage
over Carol. However, Carol is actually better than Alice at both tesuji and joseki, so Alice is
actually at a $172$ point disadvantage; the resulting $515$ point error is quite significant. The graph
of who is likely to defeat whom is not linear but circular.

It is easy to extend this example to $N$ players with $N$ attributes, where each contiguous
list of $N-1$ players (with wrap-around) has transitive ranking, but $N$ is circular.
One need only rotate the numbers 1 through $N$ on each row, to ensure that
each player is superior to the neighboring one in $N-1$ areas, similar to table  \ref{table:Circular_Ratings}.

The situation is reminiscent of a Riemann manifold, such as a sphere: while it looks planar 
in any small area, and every small area merges smoothly and continuously into the neighboring areas,
there is no way to connect them all to make one big planar surface.
Indeed, it is not even possible to distinguish a sphere from a Klein bottle based only on local information.
The same applies to table  \ref{table:Circular_Ratings} or its $N \times N$
generalization:
while every subgroup of less than $N$ adjacent players can be rated in a linear order,
and each rated subgroup merges smoothly and continuously in neighboring subgroups,
the whole group of $N$ forms a circle. In a real graph of games, it is quite likely that
some sort of high-dimensional manifold would be formed that was more complicated than 
either a line or an $N$-sphere. See section \ref{sec:MultiDimensional_Rating}.

The problem with ratings over time is that  such
a comparison is impossible without an explicit
model of what capabilities mattered, how much capability each player had in each area, and so on. 
This may be why debates about how to compare players from different eras focus precisely
on the issues what capabilities players might hypothetically have in different styles of play.
If the careers of Alice and Bob overlapped, Bob and Carol overlapped,
but Alice retired before Carol played, all we would know for certain was that Alice rated higher
than Bob and Bob rated higher than Carol. 
Because differences in odds are not always multiplicative,
the relationship between Alice and Carol would
remain unknowable unless we apply some assumption about their capability in different areas.
Particularly in the presence of statistical noise,
it is hard to determine either how large a subgroup of players is  well-approximated by a linear order
or where linearity breaks down.

Circularity is an extreme case of non-transitive ranking. It is also possible  that the ranking relations are transitive, but
not additive. Suppose Carol has an Elo advantage of 210 points over Bob, Bob has a 220 Elo point
advantage over David, but Carol only has a 270 advantage over David (not 430).
While $C \succ B$, $B \succ D$, and $B \succ D$ as transitivity requires, the Elo
advantages do not add up, so any set of $R_B$, $R_C$, and $R_D$ will have some uncertainty
when comparing any two players. This will be quantified in equation \eqref{eq:Nonlinear_Elo_errors}.

Reexamining Elo's assumption leads to some useful results. First, we go back to basics and note
that we cannot observe probability of game outcomes: we can only observe particular game outcomes.
Probabilities are inferred from observations, and frequencies are not probabilities.
For example, if A defeats B in one out of one games, it would be unreasonable to infer that
A will defeat B in 100\% future games: few would accept a 100\% probability even with
the observed 100\% frequency.
We will adopt the usual, practical methods of dealing with these problems,
even as we acknowledge that doing so evades a host of philosophical and definitional issues
which have been  discussed for hundreds of years.

A widely used method   of estimating probabilities from observed outcomes is the binomial distribution,
where the probability of success itself follows a beta distribution.
When there are $N$ wins and $M$ loses in the binomial, the posterior mean and variance of the probability's
beta distribution are given by the following standard formulas:

\begin{eqnarray}
\alpha           &=& N + 1  \nonumber \\
\beta            &=& M + 1  \nonumber \\
\mu[p]          &=&  \frac{\alpha}{\alpha+\beta}   \nonumber \\
                    &=&  \frac{N+1}{N+M+2}   \nonumber \\
\sigma^2[p]  &=&   \frac{\alpha \beta}{\alpha+\beta + 1}  \left( \frac{1}{\alpha+\beta} \right)^2
\label{eq:Binomial_Parameters}
\end{eqnarray}

The generalization to more than two outcomes is the Dirichlet distribution, which
has highly analogous results: $\alpha_i = N_i + 1$,  $\mu[p_i]  = \alpha_i / \sum_j \alpha_j$, and so on.

Notice that, for all $N, M \ge 0$,  equation \eqref{eq:Binomial_Parameters}
ensures that $0 < \mu-\sigma < \mu+\sigma < 1$.
As the Elo-advantage $A_{xy}$ is directly related to estimated probabilities, we can
use the posterior Beta distribution to get the mean and variance of $A$, similar to
what Elo provided in paragraph 2.55 of  \cite{RatingChessPlayers}.
If we calculate $A_{xy}$ from the beta distribution, then we would model it 
as a random variable with known mean and variance,  $\mu_{ij}$
and  $\sigma^2_{ij}$, for every observed pair of players $i$ and $j$.

These equations imply that there will be no problem with overshoot when
rating large tournaments. If $N$ and $M$ are both multiplied by a large
factor $k$, all that happens is that the estimated probability $\mu$
becomes closer to the observed frequency $N / (N+M)$, and the 
statistical uncertainty $\sigma^2$ declines as roughly $1/k$. 
This is in marked contrast 
to equation \eqref{eq:Classic_Elo_Update_N}.

Elo's assumption can be restated as the assumption of the existence of a set
of ranks that reproduce the observed $A_{ij}$ for $N$ players:

\begin{equation}
\forall \  1 \le i,j \le N: \ R_i - R_j = A_{ij}
\end{equation}

Because $A_{ij} + A_{ji} = 0$ and $A_{ii} = 0$, there are $N(N-1)/2$ independent
$A_{ij}$ data variables but only $N$ dependent variables for $R_i$. Thus, we have more linear equations
than free variables, so the best we can do is to find some good approximation, e.g. to minimize the expected squared error,
$S$. 

\begin{equation}
S = {\rm{E}} \left[ (R_i - (R_j + A_{ij})^2 \right]
\label{eq:Nonadditive-Error-Minimization}
\end{equation}

This is different from the system in \cite{Massey_StatModelsRanking}, which minimizes the
mean squared error in predicting points in European football. Equation \eqref{eq:Nonadditive-Error-Minimization}
minimizes the error in predicting log-odds. Not only is log-odds exactly what Elo specifies
and which is directly related to non-multiplicative odds, but it is also the relevant
quantity in actually making sports bets; see section \ref{sec:Betting}

Because there is no natural zero-point for Elo ratings, it will be mathematically convenient to let 
the ratings sum to an arbitrary value, then shift the resulting ratings
as appropriate (e.g. so that new players start at 1000).
The derivatives of equation \eqref{eq:Nonadditive-Error-Minimization}
provide a form suitable
for iterative solution. For any given set of numbers,
the value which minimizes the mean squared error is the arithmetic
average. Each line of the above system states that the $R_x$ of that line is the
arithmetic average of what would be expected from the $R_y$ and
 the advantage $A_{xy}$  for the observed $x, y$ pairs. 
 Where $N_i$ is the number of opponents for which
 $A_{ij}$ values were recorded: $N_i = |I_i|$.
 This $N_i$ is not the number of games played by agent $i$; it is
 the number of edges coming out of $i$'s node in a graph of the tournament.
 Each edge has a separate $(\alpha, \beta)$ pair summarizing
 how many wins and losses happened on that edge.
  $A_i$ is the average over $j$  of $A_{ij}$,
  we have the following condition to minimize $S$:
 
 \begin{eqnarray}
 R_{i} &=& \frac{1}{N_i} \left[ \sum_{j \in I_i}  R_{j} + A_{ij} \right]   \nonumber \\
          &=&  \left[ \frac{1}{N_i} \sum_j  R_{j}  \right]  +  \left[ \frac{1}{N_i}  \sum_j  A_{ij} \right]  \nonumber \\
          &=& \left[ \frac{1}{N_i}  \sum_j  R_{j}  \right]  +  A_i \\
\label{eq:Nonadditive-Error-Derivatives}
 \end{eqnarray}

 This can be solved as an iterative
 system, where iterations start at $t=0$.
 Unlike \eqref{eq:Consistent_Elo_Batch_Update_Algorithm}, the $A_k$ factor
 does not need to be recomputed every iteration. Again, the summation 
 is only over those $j$ indices which connect to $i$, of which there are $N_i$.
 
 \begin{eqnarray}
 R'_{t,i} &=& \frac{1}{N_i}  \sum_j  R_{t,j} + A_{i}    \nonumber \\
 R_{t+1, i} &=& \frac{ R_{t,i}  +  R'_{t,i} }{2}
 \label{eq:iterative_elo_error_min}
 \end{eqnarray}

As ${\rm{E}} \left[x^2 \right]  = \mu^2_x + \sigma^2_x$ for any random variable $x$,  we
can easily get the mean squared error between $R_i$ and any other $R_j + A_{ij}$. This
is the expected square error when we estimate the  Elo-advantage $A_{ij}$  from
fitted, linear ranks.

\begin{equation}
N_i  \sigma_i^2 =  \sum_{j=1}^{N_i}  \left[ R_i - (R_j + \mu_{ij})  \right]^2  +  \sigma_{ij}^2
\label{eq:Nonlinear_Elo_errors}
\end{equation}

Unlike previous estimates this author has found in the literature, this equation includes two sources of uncertainty:
\begin{itemize}
\item Statistical uncertainty because observed games are just small samples from a binomial process
\item Structural uncertainty because pairwise advantages may not fit exactly into a transitive, additive order.
\end{itemize}

We can separate out the two components by looking at two special cases.

If  a linear
ranking fit perfectly, then we would have $R_i = R_j + \mu_{ij}$ for all pairs.
There would still be  uncertainty due to limited sample size from binomial sampling. Therefore,
this establishes an minimum  level of statistical uncertainty:

\begin{equation}
\sigma_i^2 = \frac{1}{N_i}  \sum_{j=1}^{N_i}   \sigma_{ij}^2
\end{equation}

As usual, increasing sample sizes (e.g. playing tournaments with more games) would
increase the sample size and reduce the statistical uncertainty. 
Suppose that the statistical uncertainty were zero, $\sigma_{ij} = 0$ for all pairs. 
 If the advantages did not exactly match
the transitive, additive assumption, there
would still be structural uncertainty:

\begin{equation}
\sigma_i^2 = \frac{1}{N_i}  \sum_{j=1}^{N_i}   \left[ R_i - (R_j + \mu_{ij})  \right]^2
\end{equation}

The measurement of structural uncertainty can be done for any set of $R$ and $A$ values, regardless
of the algorithm used to produce the $R$ values.

\subsection{Uncertainty in LLS Estimates}

As discussed in section \ref{sec:Weighted_Expectations}, the least squares estimator can take
into account the relative uncertainties in ratings. In practice, this would be necessary 
because highly rated players have less uncertainty, new players have great uncertainty,
and the KR-condition is necessary, as discussed in section 
\ref{sec:Total_of_Ranks}. We treat the
prior ratings and new tournament results as uncertain estimates according to equation 
\eqref{eq:weighted_least_squared_error}. The new estimate is $R'_i$ to minimize
the weighted errors. In the right hand term, we make use of the fact
that the variance of the sum of two variables is the sum of their variances.

\begin{equation}
S =  
 \frac{\left(R'_i - R_i  \right)^2}{\sigma_i^2}
+
\sum_j  \frac{ \left[ R'_i - (R_j + A_{ij}) \right]^2}{\sigma_j^2 + \sigma_{ij}^2}
\label{eq:Weighted_Nonadditive-Error-Minimization}
\end{equation}

The solution is the following update rule.  While the expressions are long,
the estimate for $R'_i$ is just the precision-weighted average
of the old value and new observations.

\begin{eqnarray}
p_i  &=& \frac{1}{\sigma_i^2}  \nonumber \\
p_{ij}  &=& \frac{1}{\sigma_j^2 + \sigma_{ij}^2}  \nonumber \\
R'_i  &=&   \frac{p_i  R_i     +\sum_j  p_{ij} \left[ R_j + A_{ij} \right]  }  { p_i   + \sum_j p_{ij} }
\label{eq:precision_weighted_batch_update}
\end{eqnarray}

Notice that equation takes into account not only the uncertainty
in players' ratings, $\sigma_j$ but also the number of games which each pair
have played. The more games a pair plays together, the lower is the statistical
uncertainty $\sigma_{ij}$ from equation
\eqref{eq:Binomial_Parameters}. 

 If the prior uncertainty
in $\sigma_i$ is low for an experienced player, then their precision is high
and their rating
will be shifted less than that an a new player with large $\sigma_j$ and low precision.
Equation 
\eqref{eq:Nonadditive-Error-Minimization}
is just the special case of 
\eqref{eq:Weighted_Nonadditive-Error-Minimization}
when the precisions are all equal.  For the numerical examples
to follow, we will use \eqref{eq:Nonadditive-Error-Minimization}
to keep the algebra simple.

\subsubsection{Whole History Weighting}

Separating the $A_{xy}$ differences from the ratings $R_x$, without assuming
additivity, allows us to draw a close analogy to Kalman filtering, where
the $R$ are a vector of estimated state variables and $A_{xy}$ are
new observations about that state.

It is common for time-varying estimates to put exponentially decaying weights on old data.
A simple example is equation \eqref{eq:prop_control_as_long_term_average}.
Kalman filtering also has exponential weights, except that they are weighting
matrices rather than scalars.

The problem of tracking ratings over time can also be phrased as balancing the mismatch
between $R$ and $A$ with the changes in $R$. This is entirely analogous to the derivation 
of Kalman filtering, where the mismatch of state and observations is balanced against the
changes in state variables:

\begin{equation}
S = \sum_{i,t} \left[
\left(
\frac{R_{i,t} - R_{i,t-1}}{\sigma_{it}^2}
\right)^2
+
\sum_j
\left(
\frac{R_{i,t} - \left[ R_{j,t} +A_{ijt} \right]}{\sigma_{jt}^2 + \sigma_{ijt}^2}
\right)^2
\right]
\end{equation}

Again, the solution is a generalized average, except that we would obtain different
estimates for each player at each time. The equations can be arranged
as a moving average, exactly like a Kalman filter.

Exponentially decaying weights on old data can be expressed as
exponentially decaying precision in equation
\eqref{eq:precision_weighted_batch_update}. However, it would
seem to be more computationally efficient to repeatedly update
a single estimate rather than maintain a long history
(exactly the original motivation of Kalman filters).
Computationally, this means using  equation \eqref{eq:prop_control_as_average}
rather than the exactly equivalent
equation \eqref{eq:prop_control_as_long_term_average},
i. e. updating the vector of $R_i$ values as an
average of their old values and the new observations
via equation \eqref{eq:precision_weighted_batch_update}.

\subsection{Example of Non-Additive Advantage}

\begin{figure}[ht]
\begin{centering}
\includegraphics[scale=0.5]{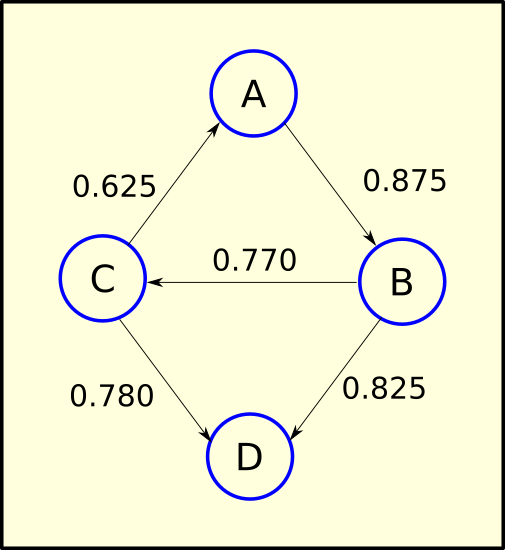} 
\par\end{centering}
\protect\caption{Nonmultiplicative Odds}
\label{fig:nonmultiplicative_probs}  
\end{figure}

\begin{figure}[ht]
\begin{centering}
\includegraphics[scale=0.5]{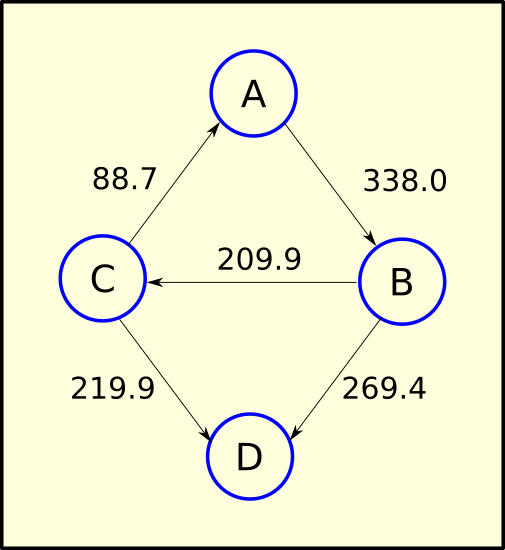}
\par\end{centering}
\protect\caption{Nonadditive Elo Advantages}
\label{fig:nonadditive_elo} 
\end{figure}

\begin{table} 
\begin{center}
\begin{tabular} {| r | r  r  r |  }
\hline
 & Base & Shifted  & Uncertainty\\
\hline
Alice   &  148.65   & 1354.65 & 218.7 \\
Bob    &  29.33   & 1235.33 & 171.2 \\
Carol  & 18.68   & 1224.68 & 171.2 \\
David  & -220.65    & 985.35 & 19.4 \\
\hline
\end{tabular}
\caption{Elo Ratings from Nonmultiplicative Odds}
\label{table:Elo_Ratings_from_Nonadditive}
\end{center}
\end{table}

\begin{figure}[ht]
\begin{centering}
\includegraphics[scale=0.5]{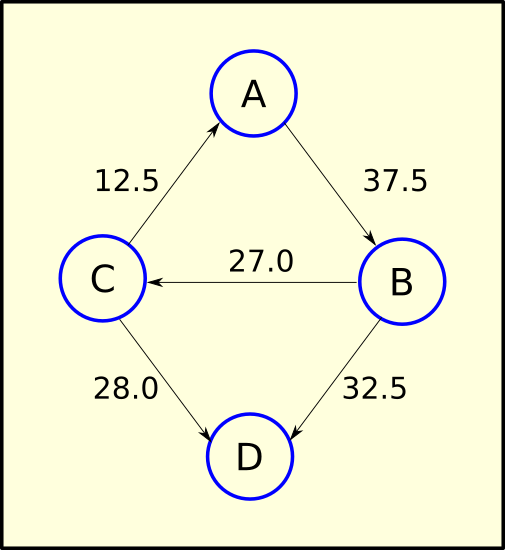}
\par\end{centering}
\protect\caption{Nonadditive English Advantages}
\label{fig:nonadditive_english}  
\end{figure}

Figure \ref{fig:nonmultiplicative_probs} presents the graph
of a tournament between four players, Alice, Bob, Carol, and David.
Each node represents a player and each line represents a set of games between that pair
of players.
While Alice probably played many games, she was only directly connected to two opponents.
To simplify the example, we depict only the directed arrows for observed pairs with $P_{xy} > 0.5$.
Notice that Bob has $3.3\!:\!1$ odds of victory over Carol, and Carol has $3.5\!:\!1$
odds of victory over David, but  Bob has only $4.7\!:\!1$ odds over David,
not $11.9\!:\!1$, which   violates  \eqref{eq:Player_Odds_Chain_Rule}.
To emphasize that the probabilities are the underlying data while rating advantages
are derived from them, we used equation \eqref{eq:Advantage_conversion} to
compute both the Elo advantages of Figure \ref{fig:nonadditive_elo} 
and the English advantages of Figure \ref{fig:nonadditive_english}.

The advantages between Alice, Bob, and Carol clearly form a cycle. Aside from the
$C \succ A$ relationship, the other four arrows do form a transitive graph, even though
they are not additive: $A_{BC} + A_{CD} = 429.8$ while $A_{BD} = 269.4$.

The earlier equations used summations and averages only over the observed
$A_{xy}$; in this example we can be more explicit about how many terms are used for each $R_i$.

 Figure \ref{fig:nonadditive_elo} shows five $A$ values which are to be approximated
 by four $R$ values. We can write the Lagrangian to minimize the sum of the squared errors,
 subject to a specified average rating $r$ as follows:
 
 \begin{eqnarray}
 \mathcal{L}(R_A, R_B, R_C, R_D, \lambda)
 &=&   \frac{1}{2} \left[ R_A - (R_B+A_{AB})   \right]^2  \nonumber  \\
 &+&   \frac{1}{2} \left[ R_A - (R_C+A_{AC})   \right]^2  \nonumber  \\
 &+&   \frac{1}{2} \left[ R_B - (R_C+A_{BC})   \right]^2  \nonumber  \\
 &+&  \frac{1}{2}  \left[ R_B - (R_D+A_{BD})   \right]^2 \nonumber   \\
 &+&  \frac{1}{2}  \left[ R_C - (R_D+A_{CD})   \right]^2  \nonumber  \\
 &-&
 \lambda(R_A + R_B + R_C + R_D -4r) 
 \label{eq:Lagranian_for_Nontransitive_ABCD}
 \end{eqnarray}

 Setting the derivatives with respect to $\lambda$ and each $R$ to zero, we get the following
 system:
 
 \begin{eqnarray}
R_A &=&  \frac{1}{2}  \left[ (R_B+A_{AB}) + (R_C + A_{AC}) +\lambda \right] \nonumber  \\  \nonumber  \\ 
R_B &=&  \frac{1}{3} \left[ (R_A+A_{BA}) + (R_C + A_{BC}) + (R_D + A_{BD}) +\lambda \right]  \nonumber  \\  \nonumber  \\ 
R_C &=&  \frac{1}{3} \left[ (R_A+A_{CA}) + (R_B + A_{CB}) + (R_D + A_{CD}) +\lambda \right]  \nonumber  \\  \nonumber  \\ 
R_D &=&  \frac{1}{2} \left[ (R_B+A_{DB}) + (R_C + A_{DC}) +\lambda \right]  \nonumber  \\  \nonumber  \\ 
4r &=& R_A + R_B + R_C + R_D 
 \end{eqnarray}

This is an example of equation \eqref{eq:Nonadditive-Error-Derivatives}
where $I_A = \{B, C\}$ so $N_A = 2$, $I_B = \{A, C, D\}$ so $N_B = 3$, and so on.
Note that once we have
any set of ranks, we can add a constant to all $R_x$ in
equation  without affecting the error terms of equation 
\eqref {eq:Lagranian_for_Nontransitive_ABCD}.
It is simplest
to set $\lambda = 0$, then shift the $R$ values to the desired mean:

 \begin{eqnarray}
R_A &=&  \frac{1}{2}  \left[ (R_B+A_{AB}) + (R_C + A_{AC}) \right] \nonumber  \\  \nonumber  \\ 
R_B &=&  \frac{1}{3} \left[ (R_A+A_{BA}) + (R_C + A_{BC}) + (R_D + A_{BD})   \right]  \nonumber  \\ \nonumber  \\ 
R_C &=&  \frac{1}{3} \left[ (R_A+A_{CA}) + (R_B + A_{CB}) + (R_D + A_{CD})   \right]  \nonumber  \\  \nonumber  \\ 
R_D &=&  \frac{1}{2} \left[ (R_B+A_{DB}) + (R_C + A_{DC})  \right] 
 \label{eq:Iteration_for_Nontransitive_ABCD}
 \end{eqnarray}
 
This is exactly what we would expect from equation \eqref{eq:iterative_elo_error_min}.

With the given data, we can solve for the first column of Table \ref{table:Elo_Ratings_from_Nonadditive}.
As $\lambda = 0$ for convenience,  the average of the base ratings is $-6.0$; we can add $1206$ to all values so as to
shift the average to $1200$. This gives the second column. We can then insert the $R$ values into equation
\eqref{eq:Nonlinear_Elo_errors}, with all $\sigma = 0$ for now, to assess
the structural 
uncertainty in predicting game-advantages. This gives the third column. Notice that while Alice,
Bob, and Carol are all involved in the non-transitive relationship at the top of Figure 
\ref{fig:nonmultiplicative_probs}, the
structural uncertainty for Bob and Carol is considerably reduced because they are also
involved in the transitive (but non-additive) relationship at the bottom of the figure.
David has the least, because he is not involved in the non-transitive cycle, but the
structural uncertainty in his rating is non-zero because the advantages
for Bob, Carol, and David are non-additive.

If we start with the unequal weights of
\eqref{eq:Weighted_Nonadditive-Error-Minimization},
rather than the equal weights of
\eqref{eq:Nonadditive-Error-Minimization},
we get a version of \eqref{eq:Iteration_for_Nontransitive_ABCD}
which uses the appropriately weighted averages.

\subsection{English Chess Federation Rating}
\label{ref:English_Rating_System}

\begin{figure}[ht]
\begin{centering}
\includegraphics[scale=0.5]{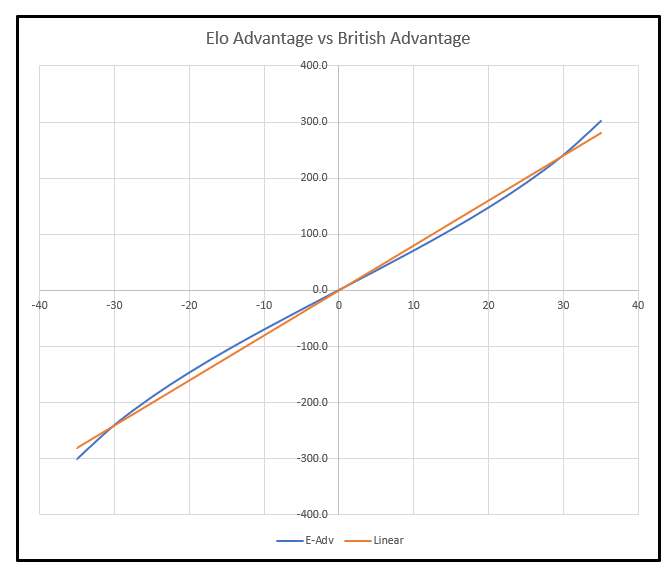}
\par\end{centering}
\protect\caption{Elo vs English Advantages}
\label{fig:elo_vs_english} 
\end{figure}

\begin{table} 
\begin{center}
\begin{tabular} {| r | r   r |  }
\hline
 & Rating  & Uncertainty\\
\hline
Alice   & 116.9 & 26.1 \\
Bob    & 105.5 & 20.8 \\
Carol  & 103.4 & 20.8  \\
David  & 74.2  & 1.2\\
\hline
\end{tabular}
\caption{ECF Ratings from Nonmultiplicative Odds}
\label{table:ECF_Ratings_from_Nonadditive}
\end{center}
\end{table}

It is useful to contrast the Elo rating system with that used by the English Chess Federation  (ECF)\footnote{As of early 2021,
the ECF website stated that
there were plans to adopt  an Elo-like system in late 2020, but
lacked a clear statement as to whether or not that change had occurred.}.
Mathematically, it seems quite different, but
in games where the players have fairly similar abilities, the rating-advantages
are highly correlated.

The ECF issues ratings to players twice a year, at the end of June and December.
The new rating is based on the average score over the last six months or 30 games, whichever is more.
Each player is awarded a score for a game based on the ECF ratings ($R_X$ and $R_Y$) of the two
players and whether $X$ gets a win, loss, or draw.
The basic rule is that $X$'s score for the game is 50 points over the opponent's rating when $X$ wins and 50 points under if $X$ loses:

\begin{equation}
S_X  =  \begin{cases}
    \max \left( R_Y + 50, R_X +10\right), &  {\rm if} \ X \succ Y \\
    \min \left( R_Y - 50, R_X - 10 \right), &  {\rm if} \  Y \succ X \\
    R_Y,                                                & {\rm if} \  {\rm{Draw}} 
  \end{cases}
  \label{eq:English_Chess_Federation_Rating}
\end{equation}

The max operation ensures that $X$ gets at least 10 points for a victory, even when $Y$ is
more than 40 points below $X$: the winner is never punished for having won.
Similarly for the min operation and a loss, which prevents
weak players from improving their rating by losing games against very strong opponents.

We can use \eqref{eq:English_Chess_Federation_Rating} to infer $P_{XY}$, assuming the difference
in ECF rating is less than 40 points. If we assume that the expected score of each player
for the game is their current rating, we can conclude the following simple formula
for the ECF advantage $R_X - R_Y$:

\begin{eqnarray}
R_X &=& P_{XY}(R_Y + 50)  +  (1-P_{XY})(R_Y - 50)  \nonumber  \\
      &=&  R_Y + 50(2P_{XY} - 1)  \nonumber 
\end{eqnarray}

Similarly,
\begin{eqnarray}
R_Y &=& R_X + 50(2(1-P_{XY}) - 1)  \nonumber  \\
      &=&  R_X + 50(1 -2P_{XY})  \nonumber 
\end{eqnarray}

Taking the difference and dividing by two, we get the elegant formulas converting 
between probability of victory and the ECF advantage:

\begin{eqnarray}
R_X - R_Y &=&  100 \left( P_{XY} - \frac{1}{2} \right)  \nonumber \\
P_{XY} &=& \frac{1}{2} + \frac{R_X - R_Y}{100}
\end{eqnarray}

If we denote the Elo advantage by $A$ and the ECF advantage by $B$, we reach a simple
conversion formula for advantages in terms of victory probability $p$.
Again, this applies when $|B| < 40$ and $|A| < 382$, i.e. $0.1 < p < 0.9$.

\begin{equation}
\frac{50 - B}{50 + B}  \ = \frac{1-p}{p}  \ = e^{+\beta A}
\label{eq:Advantage_conversion}
\end{equation}

As $B > 40$ implies an estimated win probability over 90\%,
which is an advantage of over 382 Elo points,
it would seem rather unsporting to play games where $|B| > 30$.
Equation \eqref{eq:Advantage_conversion} specifies a relationship
between $A$ and $B$ which is clearly nonlinear, but in the range
of $|B| \le 35$, it is surprisingly close to straight line as illustrated 
in Figure \ref{fig:elo_vs_english}.
Several different fitting procedures give almost exactly the same relationship
between ECF advantages and Elo advantages:
\begin{equation}
A \approx 8 B
\end{equation}

The relationship is almost exact when $B \approx  30$ and $A \approx 240$.
We could expect this relationship to hold closely if the underlying probabilities
met the assumptions of both models, which our example data does not.
As of early 2021, the ECF was considering a change to an Elo-like system.
The new ratings were to be specified by a linear conversion: $R' = 7.5 \times R + 700$.
This would imply that a factor of $7.5$ between ECF advantages and Elo advantages
better reflects the various differences between the real-world data and the
theoretical systems.
 
The Elo and ECF systems diverge when considering games between very dissimilar
players. If $X$ has 95\% chance of defeating $Y$, the Elo advantage is about
552 points, but equation \eqref{eq:Advantage_conversion} is not valid
for $p > 0.9$ The ECF system does not directly provide a way to estimate
the win probability when one player has more than an 80-90\% chance
of defeating the other.

Given the win-probabilities of figure \ref{fig:nonmultiplicative_probs},
we can write the equations stating that the expected scores in the ECF
system match the current ratings. Assuming that the marked arrows are 
played with equal probability,

\begin{eqnarray}
R_A &=& \frac{1}{2} \left[ 
R_B + (2P_{AB} -1)  + R_C + (2P_{AC} - 1)
  \right]   \nonumber  \\
R_B &=& \frac{1}{3} \left[ 
R_A + (2P_{BA} -1)  + R_C + (2P_{BC} - 1) + R_D + (2P_{BD} - 1)
  \right]   \nonumber  \\
R_C &=& \frac{1}{3} \left[ 
R_A + (2P_{CA} -1)  + R_B + (2P_{CB} - 1) + R_D + (2P_{CD} - 1)
  \right]   \nonumber  \\
R_D &=& \frac{1}{2} \left[ 
R_B + (2P_{DB} -1)  + R_C + (2P_{DC} - 1)
  \right]  
\end{eqnarray}

Assuming an average ECF rating of 100,
we can solve for the first column of Table \ref{table:ECF_Ratings_from_Nonadditive}.
Similar to the Elo case, we can look at the RMS error to get the second column.
As would be desirable, the ratings of the players are in the same order, and the
magnitudes of uncertainties follow the same pattern.

The Elo and ECF ratings show about 99.8\% correlation, where the deviations from
their respective means have about the expected proportionality:

\begin{equation}
{\rm{ELO}}_X - 1200  = 8.48 (  {\rm{ECF}}_X - 100 )
\end{equation}

\subsection{Comparison of Elo Rating Algorithms}
\label{sec:Multi_Elo_Alg}

\begin{figure}[ht]
\begin{centering}
\includegraphics[scale=0.5]{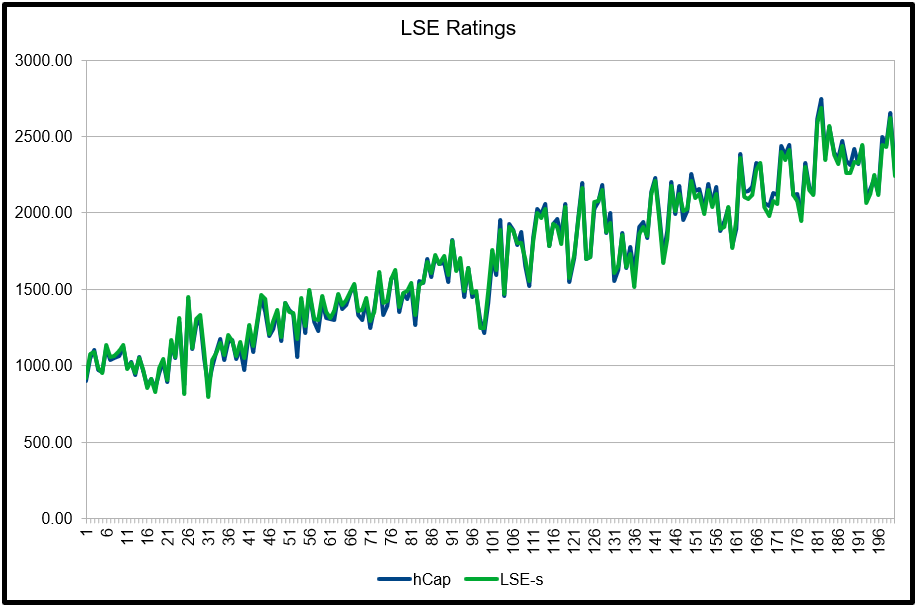} %% this is the name of the image file
\par\end{centering}
\protect\caption{True capability and LLS-fitted Elo ratings}
\label{fig:LSE-fitted-ranks}  %% label by which it is referenced 
\end{figure}

\begin{figure}[ht]
\begin{centering}
\includegraphics[scale=0.5]{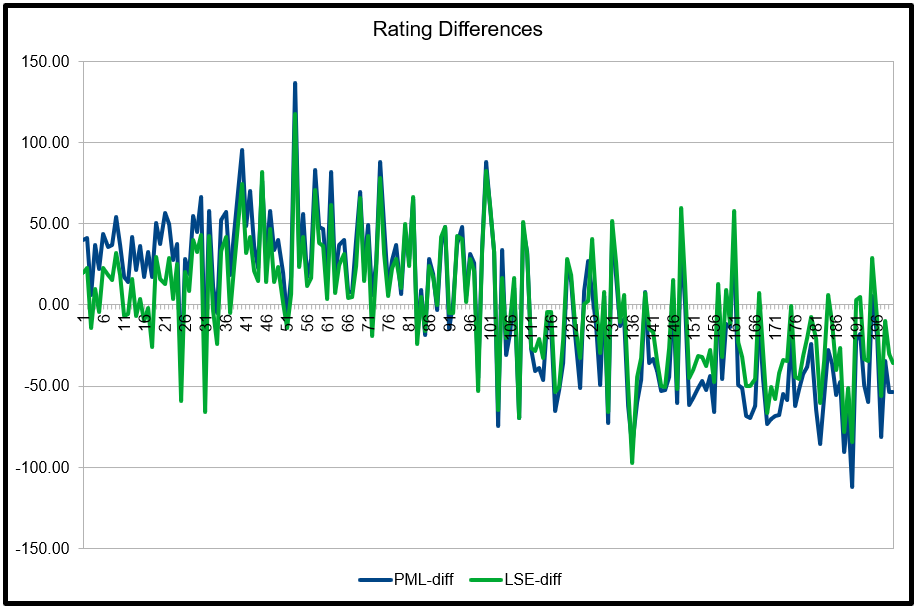} %% this is the name of the image file
\par\end{centering}
\protect\caption{Difference in ratings estimated by PML and LLS}
\label{fig:PML-LSE-errors}  %% label by which it is referenced 
\end{figure}

Notice that we now have two very different
ways to calculate Elo ratings.
First, there is the posterior maximum likelihood  (PML) estimator
\eqref{eq:Vector_SC_Elo}
as calculated by the algorithm in
\eqref{eq:Consistent_Elo_Batch_Update_Algorithm}.
This requires a Bayesian prior estimate
for the ratings, which may be
an uninformative prior if desirable.

Second, there is
the linear least squares (LLS)  error  estimator
\eqref{eq:Nonadditive-Error-Minimization}
as calculated by the algorithm in
\eqref{eq:iterative_elo_error_min}. This
does not use a Bayesian prior for the ratings,
although it does use a Bayesian prior for the win
probabilities. Unlike the PML method,
equation  \eqref{eq:Nonlinear_Elo_errors}
provides a very simple way to include uncertainty
from both the small-sample effect and
the structural uncertainty from non-multiplicative probabilities.

The classic Elo system suffers from overshoot for large tournaments.
The minimum squared error estimator does not: large
tournaments  result only in the statistical uncertainty
becoming smaller.

There are significant mathematical differences between the LLS and PML estimators,
but the results are largely the same numerically.
Figures \ref{fig:PML-fitted-ranks}, \ref{fig:LSE-fitted-ranks} and  \ref{fig:PML-LSE-errors} illustrate the following points,
where the same dataset was given to both algorithms.
\begin{enumerate}

\item Convergence
\begin{enumerate}
\item  For the same data set, the LLS algorithm typically takes one quarter to  one tenth as
many iterations to reach the same level of precision as the PML algorithm
(decimals after convergence).

\item Each iteration of the LLS algorithm is faster than each iteration of PML.
Each iteration of LLS is just a simple average. For the PML, the new expected win
rate for each player must be recalulated from the latest estimated ratings of the other
players at each iteration.

\item For the same data set and the same precision, the PML and LLS algorithm
consistently produce almost exactly the same answer. For this data set, the
correlation of the PML and LLS estimates was slightly over 99.99\%.
Figures  \ref{fig:PML-fitted-ranks} and \ref{fig:LSE-fitted-ranks} would have been
very difficult to distinguish if they had used the same color scheme.
Figure  \ref{fig:PML-LSE-errors} shows ``detrended'' data by subtracting
the estimated capability from the PML and from the LLS estimates. The
blue and green lines are both above zero when both estimators were high,
both below zero when both estimators were low, and so on.
The differences had about 94.3\% correlation, because statistical variations in the 
game outcomes affected both equally. 
The RMS difference of the PML is 46.1
Elo points, while the LLS was about 41.1, which is not a significant difference.

\item For this particular dataset, there is some tendency in figure  \ref{fig:PML-LSE-errors} for the PML estimate
to be slightly higher than the LLS for low capabilities and slightly lower for high capabilities.
However, it is less than the random variation of either.

\item Both FIDE for Chess and EGF for Go avoid the classic Elo update formula
in the case where a player has  won  (or lost) all the games. The reason
is that a 100\% win (or loss) rate can only be obtained by an infinitely
high (or low) Elo rating. The LLS estimator does not suffer from this problem,
even in the extreme case of a single game, so no special formulas or
procedures are necessary.
\end{enumerate}

\item For small numbers of games, the LLS algorithm consistently produces
smaller differences in players' ratings. For example, if a strong player defeats a weak player
in two out of three games, the LLS algorithm uses the Beta distribution to estimate
a probability of $(2+1)/(3+2) = 0.600$ (about 70 Elo points) while the PML estimates
$2/3 = 0.667$ probability  (about 120 Elo points). The relative advantages are simply
scaled down because LLS takes more games to prove a high win rate; whether this is
a bug or a feature might be debated.
The deviations of the LLS ratings from their average is almost perfectly correlated with the deviations
of the PML ratings from their average (routinely over 99\%).
The contraction toward
the mean is a simple rescaling that affects all players equally, without affecting their
relative ratings at all.
      \item The contraction can be largely eliminated by reducing the prior
weight in the beta distribution from $1.0$ to something like $0.1$. This brings the
estimated probabilities much closer to the observed frequencies. For winning two
out of three games, this would shift the observed probability to $(2+0.1)/(3+0.2) =
 0.656$, which is about 112 Elo points. The lower prior weight was used to produce
 figure \ref{fig:LSE-fitted-ranks} which is why it shows no contraction to the mean.
\end{enumerate}

While both algorithms are analytically correct and yield equivalent results, the LLS algorithm 
appears to have superior numerical performance. However, small numbers of games
(less than a hundred for each player) will inherently have the statistical variations
pointed out by Elo in figure \ref{fig:elo-reliability}, with consequent high
uncertainty in Elo points, regardless of what rating algorithm is used.

\section{Multi-Dimensional Advantage}
\label{sec:MultiDimensional_Rating}

Notice that if we change the 1-dimensional scalar ranks, $R_i$ to $K$ dimensional points $P_i$ on a manifold,
then the optimization problem becomes one of finding points such that the distance between them
matches the observed Elo-advantages. If each $A_{ij} > 0$ defines a directed line segment,
what is the minimum number of dimensions to embed the graph?
For the Alice, Bob, Carol example above, we could obviously achieve a perfect
fit in two dimensions by any equilateral triangle with directed edges of length 172:
one dimension is not enough but two will suffice.

\subsubsection{Well Ordered Clusters}

A modified version of $K$-means could used to find clusters 
so that the total ordering error was minimized. That is,
$i$ and $j$ are well-ordered if the mismatch of $R_i -R_j$ and
$A_{ij}$ is small compared to the uncertainty in $A_{ij}$.
We denote this by $C(i,j,k)$ for distinct indices:

\begin{equation}
C(i,j,k) =\frac{
\left( A_{ij} + A_{jk} - A_{ik} \right)^2
}
{
\sigma^2_{ij} + \sigma^2_{jk}  + \sigma^2_{ik} 
}
\end{equation}

For a given group of points, $G$,  can define the measure  $m(G)$
as the maximum of $C(i,j,k)$ for distinct indices in $G$.
Analogous to $K$-means, we could randomly assign points to
groups  then start shifting them to reduce the RMS or maximum
of $m(G)$ until no more progress could be made.

If we define $c$-consistency of a group to mean $m(G) \le c^2$, then
several questions immediately arise:
\begin{enumerate}
\item For a given set of $A$ and a given $c > 0$, how many different $c$-consistent groups exist?
\item How much do they overlap?
\end{enumerate}

\subsubsection{Separating Hyperplanes}

The hyperplane through $P_i$ perpendicular to the segment
from $P_i$ to $P_j$  is the set of $Q$ 

\begin{equation}
H_{ij} = \{ Q \ | \ (Q-P_i) \cdot (P_j - P_i) = 0 \}
\end{equation}

This defines a half-space of points with positive value:

\begin{equation}
S_{ij} = \{ X \ | \ (X-P_i) \cdot (P_j - P_i) > 0 \}
\end{equation}

Obviously, $P_j$ is in the positive half-space $S_{ij}$. Because $A_{ij} + A_{ji} = 0$,
we can, without loss of generality, consider only those $i,j$ for which $A_{ij} > 0$.
The hyperplane $H$ defines a local rating scale, where point $P_i$ is rated
higher than point $P_j$.

We want the spacing of $P_i$ and $P_j$ to correspond to the difference $A_{ij}$, while
respecting the sign.  If it were possible to assign $K$-dimensional locations to all $P_i$
such that the following held for all $i, j$ where $A_{ij} > 0$, then we would have a
perfect fit:

\begin{equation}
A^2_{ij} =  (P_j -P_i) \cdot (P_j - P_i)  
\end{equation}

One obvious algorithm is to generate the initial estimates $P_i$ at random, then incrementally
shift each to reduce its error, while keeping the cloud centered on the origin.
In this way, we could begin to look for subsets that were correctly ordered by a single hyperplane.
Finding the minimum number of hyperplanes such that every pair was correctly ordered
by at least one hyperplane would probably reveal underlying structure.

\subsection{Two-Factor Elo Parameters}

In \cite{Kiraly_ModellingCompetitiveSports} several ways of rating
competitive team sports teams, specifically the 47 teams of the English Premier League.
The mathematical foundations of the classic Elo system are reviewed, as well as corrections
to several misconceptions that have grown up around the system since Elo introduced and
explained it. Alternative rating systems are analyzed; we discuss some but not all.
The Elo system was developed for games like Chess or Go, where there are simply win,
lose or draw outcomes. In games like football or basketball, the winner is determined
by who gets the most points, and large margins of victory are important for betting.
The final, and most detailed,  analysis of the paper uses a non-Elo system to predict the outcome by
points.

First, they present the classic Elo system as a one-factor model,
where $h$ is the home-team advantage:

\begin{eqnarray}
A_{ij} &=& Z_{ij} + h_{ij} + \eta_{ij}  \nonumber \\
Z_{ij} &=&   R_i - R_j  
\label{eq:one_factor_Elo}
\end{eqnarray}

They mention that this can be extended to a two-factor model:

\begin{eqnarray}
A_{ij} &=& Z_{ij} + h_{ij} + \eta_{ij}  \nonumber \\
Z_{ij} &=&   u_i v_j - u_j v_j  
\label{eq:two_factor_Elo}
\end{eqnarray}

Of course, with $2N$ parameters to fit the data instead of just $N$, the RMS error falls,
but not greatly. The cost is that there is no simple way to incrementally update
the $(u,v)$ pair for a player. This author did calculate the PML estimate for $u$, and it
is essentially the same as the SC-Elo except that the wins and expected 
wins are unevenly weighted by the opposing $v$.

As we have mentioned, the $Z$ of equation \eqref{eq:one_factor_Elo}
is not altered if a constant is added to both $R$ terms on the right side. However,
this is not true of equation \eqref{eq:two_factor_Elo}: neither $u$ nor $v$ can be
shifted without changing the left side. 

\begin{eqnarray}
Z'_{ij} &=&  \left(  u_i +a \right) \left( v_j + b \right)  -   \left(  u_j +a \right) \left( v_i  + b \right)   \nonumber \\
          &=&  Z_{ij} + a  \left( v_j  - v_i \right) + b \left(  u_i - u_j  \right) 
\end{eqnarray}
 
Note that the effect of $a$ cancels when $v_i = v$.
The one-factor model is a special case of the two-factor, with $R=u$ and $v = 1$, which is why the
one-factor model can have $R$ arbitrarily shifted. This no longer happens in the two-factor model,
which has natural zero points for both $u$ and $v$, and thus
inherently has both positive and negative values for $u$.

However, we could rescale them in complementary
ways:

\begin{eqnarray}
Z'_{ij} &=&  \left( a  u_i   \right) \left( b v_j  \right)  -   \left(a  u_j  \right) \left(b  v_i ) \right)  \nonumber \\
          &=& a b  Z_{ij} 
\end{eqnarray} 

There is obviously no change if $ab=1$. 

 For brevity,
we will write dot products as $\sum_i u_i v_i = uv$,  $\sum_{ij} A^2_{ij} = AA$,
 $\sum_{ij} Z_{ij} \eta_{ij} = Z\eta$,
and so on. As mentioned, even if we  rescale $v$ so that $v  v = 1$,
so that the $u$ values have the same magnitude as the $R$ values,
the $u$ values will turn out both negative and positive.

The scale of $u$ and $v$ is jointly set by $A$.  If we assume $h=0$  temporarily
and that  $Z$ is the optimal fit in 
equation \eqref{eq:two_factor_Elo}, so that $Z  \eta = 0$, we get
the following:

\begin{equation}
AA = 
2\left[
(uu)(vv) - (uv) (uv)
\right]
+
\eta  \eta
\end{equation}

The $\eta \eta$ term is just the sum of the squared errors, so it is always non-negative.
The term in square brackets is also non-negative by the Cauchy-Schwarz in equality,
so both sides are non-negative as expected. Because the $AA$ term is a constant,
minimizing the error term $\eta \eta$  should make $u$ and $v$ nearly perpendicular
(which does not drive $Z$ to zero).

Secondly, the discussion in  \cite{Kiraly_ModellingCompetitiveSports} also
considers several models with ternary outcomes, i.e. $\{ \rm{win},  \rm{draw},  \rm{loss} \}$.
The model first estimates the log-odds of each independently, which does not even guarantee that
the probabilities sum to one. Presumably, in practice the optimal estimates
would add to nearly one, then be  rescaled to exactly one,
but then they are no longer the optimal estimates (which did not add to one).
When large quantities of money are bet, or franchises valued, and so on, it would
be important to preserve optimality.

The second model fits two parameters, $L_{ij}$ and $\phi_{ij}$,
when team $i$ faces team $j$,  using the familiar sigmoid function, $\sigma(x)$

\begin{eqnarray}
\sigma(x) &=& (1 + e^{-x})^{-1} \nonumber  \\
\sigma(x) + \sigma(-x)  &=&1 \nonumber  \\
\lim_{x \rightarrow +\infty} \sigma(x) &=& 1 \nonumber  \\
\lim_{x \rightarrow -\infty} \sigma(x) &=& 0  
\end{eqnarray}

They estimate the probabilities as follows: 
\begin{eqnarray}
W_{ij} &=& \sigma(L) \nonumber  \\
L_{ij}  &=& \sigma(-L-\phi) \nonumber  \\
D_{ij}  &=& 1 - (W_{ij}  + L_{ij} )
\label{eq:second_Kiraly_model}
\end{eqnarray}

A problem with this model is that if $L$ is large, $W \approx 1$. If $\phi = -2L$, then
$L \approx 1$, so $D \approx -1$. Because Bayes' Theorem requires integrating over the 
full range of parameters, this presents a problem: we have to integrate over negative probabilities,
even though they are likely to have a very low prior probability. Numerically, we can get around this
by limiting the range of integration, effectively setting the prior to zero for $(L, \phi)$ combinations
which yield invalid results for the given data set.
However, this puts us in the odd position of having to inspect the data  to determine  {\it a posteriori} what
should be  our {\it a priori} beliefs.  Like the first model, the fact that it is even possible to
generate invalid probabilities shows that there is a fundamental problem with the model.

The problem is easily fixed. This author proposes the following two parameters.

\begin{eqnarray}
\alpha_{ij} & =& \ln \left[ \frac{W_{ij}}{L_{ij}}\right] \nonumber  \\
\beta_{ij} & = & \ln \left[ \frac{D_{ij}}{1 - D_{ij}}\right] 
\end{eqnarray}

It is easy to see that any $(\alpha, \beta)$ pair yields estimates consistent with the constraints that  $W +D+L=1$
and $0 \le W, D, L$.

The discussion in   \cite{Kiraly_ModellingCompetitiveSports} 
uses the frequentist or maximum likelihood estimates $p = N/M$ for the
estimated probability of winning if the team won $N$ out of $M$ games,
and similarly for the probabilities of loss or draw. For a whole
season of play, this presents no problem, as every team is likely to have both wins,
draws and losses. However, in the context of small tournaments, like updating
an Elo rating after just a dozen games or so, it is quite possible
that a player may have won none (or all) the games, lost all (or none), or drawn all (or none).
Anatoly Karpov defended his chess world champion title against Viktor Korchnoi
in  1978 with 6 wins, 5 loses, and 21 draws, then again in 1981 with 6 wins, 2 loses, and 10 draws.
In bouts against Garry Kasparov in 1984, 1985, 1986, 1987 and again in 1990,
Karpov achieved 19 wins, 21 loses, and 104 draws.
Thus, substantial sequences of drawn games are not implausible even between the same two players.
In 1907, Edward Lasker achieved eight wins, seven draws, and zero losses against Frank Marshall.
In 1921, Capablanca defeated Lasker with four wins, zero losses, and 10 draws.
If we treated observed frequencies as probabilities,
both these contests would create problems with either infinite odds or logarithm of zero odds.

In equation \eqref{eq:second_Kiraly_model}, setting any of the
$W$, $L$, or $D$ values to observed frequencies  of either zero or one gives numerically
unreasonable results. The log-odds go to plus or minus infinity, the estimated ratings go
to plus or minus infinity, squared-error estimators fail, and so on.

To avoid the problem of probabilities being either 1 or 0, we recommend using the 
Dirichlet distribution for the posterior probability estimates based on observed outcomes:

\begin{eqnarray}
N &=& N_W + N_D + N_L  \nonumber  \\
W & =& \frac{N_W + 1}{N+3} \nonumber  \\
D & =& \frac{N_D + 1}{N+3} \nonumber  \\
L & =& \frac{N_L + 1}{N+3} \nonumber  \\
\frac{W}{L} &=& \frac{N_W + 1}{N_L + 1} \nonumber  \\
\frac{D}{1-D} &=& \frac{N_D + 1}{N_W + N_L + 2}
\end{eqnarray}

Fourthly,  \cite{Kiraly_ModellingCompetitiveSports}  focuses on 
the margin of victory, as an integral number of points. For sports betting,
this is far more useful than a simple win probability, as bets are often made
on whether a team can exceed the point spread, not just win or lose.
Most English football matches score
in the single digits, unlike American basketball, so the integer nature of
football scores is important. The probability of a win
is then simply the probability of a positive margin, draws are a zero margin,
and losing is a negative margin.

As it is not an Elo-type system, we do not discuss their model of margin of victory.

\clearpage
\section{Asymmetric Games}
\label{sec:Asymmetric-Games-Rating-ANOVA}

\subsection{Battle for Moscow}

\begin{figure}[ht]
\begin{centering}
\includegraphics[scale=1.5]{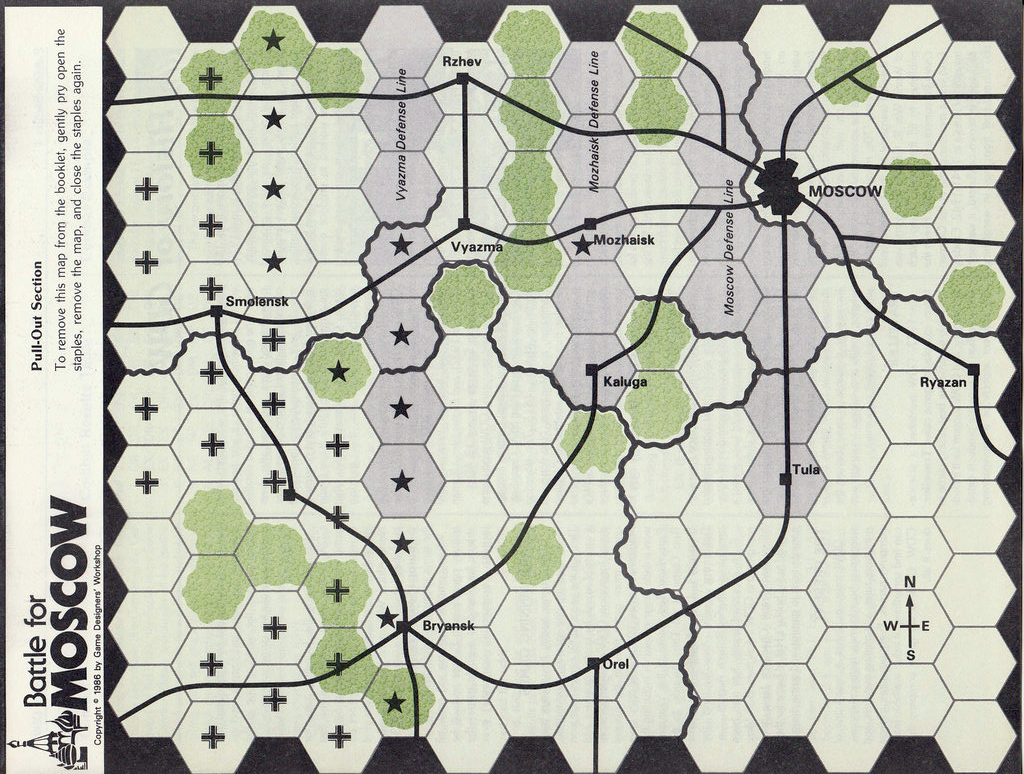}
\par\end{centering}
\protect\caption{Battle for Moscow Map}
\label{fig:battle_for_moscow} 
\end{figure}

Games like Chess and Go are essentially symmetrical between Black and White.
Games like Football are nominally symmetric, but the home-field advantage is
a well-known factor. In the Elo-like scheme described in \cite{RankingSystemsInFootball},
the home-field advantage is represented by adding exactly 100 Elo-points to the 
home team's rating. It would be useful to determine a value from observations
rather than postulating a number, and equation \eqref{eq:side_advantage} does so.

Commercial strategy games, such as the 1970's style paper strategy games on hexes, are often quite
asymmetric between the players. The map for a very small\footnote{This map
is 14 by 10, totalling 140 hexes, with a few dozen ground units.
Panzergruppe Guderian \cite{PGG76} is a fairly standard
game whose computational complexity dwarfs that of Go. The map is 59 by 31, giving
1829 hexes of open, forested, swamp, water and  city terrain,
crossed by lines of  road, railroad and river. About 250 units occupy
the map and any number can be moved in one turn by each player.}
 introductory game \cite{BFM76}
is shown in Figure \ref{fig:battle_for_moscow}.
The task of the German player is to start
at the hexes marked with Iron Cross symbols, then move east and occupy
Moscow. The Russian player starts with units in a line above Bryansk marked
by stars. In this game, the starting positions are always the same, but it
is common in other strategy games for each side to setup their
pieces relatively freely within the constraints of historical plausibility.
Both sides can bring a limited number of reinforcements from
their edge of the map. The game runs
seven turns, and whoever occupies Moscow at the end wins.
The most fundamental difference is that one side plays offense and must
change the situation to win while the other plays defense and need
merely preserve the status quo (i.e. continue holding Moscow). 

Moreover, each side has different kinds of units, with different ways of moving. 
Most obviously, the initial placement of units is quite asymmetric: German
units start in a dense mass in the western third of the map, while
the Russian units start in a thin line also in the western third. The entire
eastern 2/3 of the map (including Moscow) starts as Russian-held territory
but without units occupying it.
The Russian units can move swiftly along the smooth lines representing
railroads, but the German units cannot: German trains had their wheels
too close together to use Russian tracks.
On the other hand, the German units are much more mobile cross country.
The wiggly lines between hexes represent
rivers, which are difficult for each side to cross.
Each type of hex
affects the mobility and combat power of units in it, and the hex-types are
not distributed uniformly. There are various other
differences between the two sides.

In terms of training
artificial intelligence to recognize and respond to visual patterns, the two sides
are quite different, so we might well expect a software agent to earn
one Elo rating as German and a different one as Russian. Indeed, it may turn
out that the game is ``unbalanced'', in that one side in general has an advantage
-- exactly like the home team advantage in football. A major design goal
of recreational strategy games is that they should be balanced despite being asymmetric.
We will refer to paper strategy games generically, but examples like
of \cite{PGG76} and \cite{BFM76} should be kept in mind.

\subsection{ANOVA for Asymmetric Tournaments}
\label{sec:Asymmetric-Tournament-ANOVA}

\begin{figure}[ht]
\begin{centering}
\includegraphics[scale=1.0]{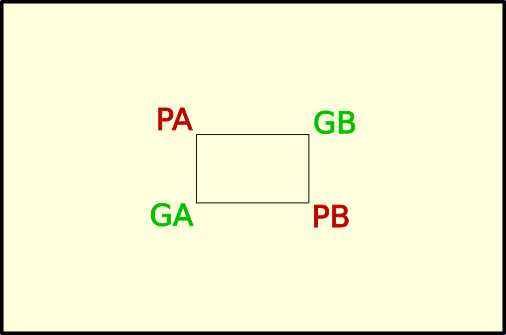}
\par\end{centering}
\protect\caption{Two Agent Asymmetric Tournament}
\label{fig:two_agent_colored_tournament} 
\end{figure}

To introduce terminology, we will introduce two generic sides, Pink and Green.
They could represent Black and White in Chess, German and Russian in Battle for Moscow,
home team and visitor in Football, and so on. The non-standard names are chosen
to avoid any association with standard tactics expected for any particular side. For example,
Black in Go is always the weaker player, always moves first, potentially with several
sequential moves as a handicap,
so the strategy and tactics of Black can differ substantially from White. 
Similarly, Blue in the Cold War represented NATO while Red represented the
Warsaw Pact, and there were well known tactical and technological differences between the
two sides.

We will assign two Elo ratings to each agent: one rating for play as Pink and one as Green.
To do this, we extend equation \eqref{eq:Nonadditive-Error-Minimization} by
replacing $R_i$ with the pair $(RP_i, RG_i)$

\begin{eqnarray}
RP_i - RG_j  &=& DPG_{ij}  \nonumber  \\
RG_i - RP_j  &=& DGP_{ij}  
\end{eqnarray}

Notice that, because $RP_i$ and $RG_i$ are separate ratings, there is no special
relationship between $DPG_{ij}$ and $DGP_{ij}$. However, $DPG_{ij} + DGP_{ji} = 0$.
Algebraically, this simply replaces LLS estimation with $N$ ratings by the 
exact same system with $2N$ ratings. No special solution routine is required,
though the wins and losses of an agent playing as Pink must be tracked separately from the
same agent as Green to get the input data right.

Figure  \ref{fig:two_agent_colored_tournament}  shows that the Pink and Green ratings
do not separate out into two different systems. The graph depicts a round-robin tournament
between agents A and B. On the top edge, A plays Pink against B as Green, probably
a few dozen times for software agents.
On the left edge, A plays Pink against A as Green, multiple times. The other edges
are similar. Notice that we can compare the ratings of A as Pink with the
rating of B as Pink to determine which plays that role more skillfully,
but we can only have contests
between Pink and Green.
Because the LLS fitting solves
for all four $R$ as one linear system based on the four independent $A$ values,
each ratings affects the others. For example,
we could raise all four by 100 Elo points with no net effect, but we could not
raise just the Pink ratings by 100. Following standard ANOVA practice,
we would subtract off the mean so that the sum of all Pink and Green ratings
for all agents is zero.

\begin{figure}[ht]
\begin{centering}
\includegraphics[scale=1.0]{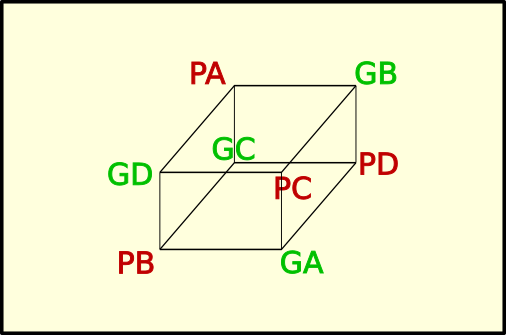}
\par\end{centering}
\protect\caption{Four Agent Asymmetric Tournament}
\label{fig:four_agent_colored_tournament} 
\end{figure}

Figure  \ref{fig:four_agent_colored_tournament}  shows a similar tournament
with four players, arranged at the vertices of a cube. For clarity, the four diagonal lines for self-play
are omitted, but the $PA\!:\!GA$ through $PD\!:\!GD$ lines could be included if desired.
Again, all eight ratings are interrelated, and changing any one of the twelve independent
$A$ values will affect
several Pink and several Green ratings.
For simplicity, we have presented a balanced design so that the ANOVA is equally simple.
All the usual machinery, such of balanced factorial designs, could be employed;
many textbooks on those topics are readily available.

As is typical with ANOVA, we can determine the overall Pink vs Green imbalance
as either the difference between average ratings or as the average of the 
difference between ratings:

\begin{eqnarray}
RP &=& \frac{1}{N} \sum_{i=1}^N RP_i \nonumber  \\
RG &=& \frac{1}{N} \sum_{i=1}^N RG_i \nonumber  \\
\rho &=& \frac{RP  - RG }{2} 
\label{eq:side_advantage}
\end{eqnarray}

Note that $RP + RG = 0$, because the sum of the $RG_i$ and $RP_i$ is zero.
This $\rho$ is the handicap to make $RP - \rho = RG +\rho$.
It  represents the Elo-advantage from playing Pink,
which could represent the home team advantage, the advantage
of  Pink over  Green (or vice versa), and so on.

Finally, we can determine the overall rating for each agent:

\begin{eqnarray}
R_i  &=& \frac{1}{2} \left[ (RP_i - (RP - \rho)) + (RG_i - (RG + \rho))  \right]   \nonumber  \\
     &=& \frac{1}{2}  (RP_i + RG_i)  - \frac{1}{2} ( RP + RG +\rho - \rho)    \nonumber  \\
     &=& \frac{1}{2}  (RP_i + RG_i)  
\end{eqnarray}

As typical ANOVA, the
idiosyncratic $\rho_i$ is the residual not accounted for
by the overal $\rho$.

\begin{eqnarray}
RP_i - (\rho + \rho_i)  &=&  RG_i + (\rho + \rho_i)   \nonumber   \\
RP_i - RG_i  &=&  2(\rho + \rho_i)   \nonumber   \\
\rho_i   &=&   \frac{(RP_i - \rho)  - (RG_i  + \rho) }{2} 
\end{eqnarray}

\subsection{Tournament for Evaluating AI}

In most experiments with machine-learning in competitive games, an initial random agent ($A_0$) is generated.
Then more agents are created and trained against $A_0$ until one performs significantly better (e.g. at least 55\% out of 400).
It then becomes $A_1$, and more agents are created and trained against both $A_0$ and $A_1$ until $A_2$ is found,
and so on. This creates a ``ladder'' of progressively stronger agents, each developed to defeat its predecessors.
Two lessons from Alpha Zero may be relevant.
First, training is limited to a mix of fairly recent opponents.
It would be an uninformative waste of time to play vastly weaker opponents, and playing against only the most recent might
lose the ability to defeat a variety of opponents. Second, again to ensure that it can cope with a variety of styles,
the move selection should be stochastic, not deterministic.
The Japanese rules for Go specify that Black must always setup the initial stones on the star points,
while the Chinese rules allow Black to place the initial handicap stones anywhere.
With commerical paper  strategy games, the initial setup should also be varied. 
This ensures that there is always variety and 
unexpected moves and hence that the AI agents do not develop a pathological style of play that only works
against copies of themselves.

We can follow the same basic process, adjusted to take into account the fact most paper strategy games are asymmetric
between between two sides. Learning to play several games would involve multiple terrains, and which could use different criteria,
so the ANOVA would have to be extended in the obvious way to take into account more factors.

For training on a specific game like \cite{BFM76}, the most basic agent, $A_0$, would be not a random neural network but
a pair of C2 plans for Pink and for Green; each
should have  basic branches and sequels. Like beginners in chess, $A_0$ would be assigned a fixed rating of 1000, because
it can never improve or learn. The next agent, $A_1$ could be trained against $A_0$, being careful to compare their
performance only when they are in similar roles, in similar situations, etc. as discussed above. Once $A_1$ is shown to be
significantly better, e.g. a 50-point Elo improvement in 400 non-interactive competitions, it would be ``frozen'' and
not changed anymore, with its rating permanently fixed. As many training techniques use a database of past situations
for ``experience replay'', this would be achieved simply by preserving records of past play without alteration.
 
Because human patience is not a factor in AI-vs-AI play, it is practical to use large
tournaments (e.g. 400 games for Alpha Zero), in order to reliably assess even small advantages, like the 34 Elo-point
advantage implied by a 55\% win rate. The above design of four strategy games could be expanded to 400 simply
by slightly varying the initial conditions of each scenario to give 100 variants on each.
 
Note that the general concept to training AI via  self-play requires that each 
agent be able to play both Pink and Green.\footnote{There are many proposed schemes
for self-play in training AI, with widely varying names and approaches. The requirement to play both
sides applies to all of them.}
For Alpha Zero, playing both sides of Chess, Go, or Shoji was simple because the boards, 
pieces, and rules are symmetric between the two sides. This is not true in strategy games, where
the two sides may have different types of forces, starting at different sides of an irregular terrain.
As the $A_i$ agents   get better, they cannot continue simply playing the Pink side
against the same basic Green from $A_0$: they would  learn to beat the
basic, non-adaptive Green agent  and then stop learning because there was no further challenge.
It is essential to the learning process
that it keep playing against better and better versions of itself, which requires playing both Pink and Green.

\clearpage
\section{Margin of Victory}
\label{sec:Margin_of_Victory}

In many commercial strategy games, it is a design goal to reflect the fact that, in real combat, the margin of victory
is important: a decisive victory is much better than just barely winning at all.
This may reflect the fact that real combat outcomes are stochastic, 
  so a wider ``margin of safety'' is better.
Equation \eqref{eq:Bradley_Terry} is the foundation of the Elo system, but it has no way of distinguishing
between a comparison which was just barely won and one which was decided by an overwhelming margin.
There are at least two ways to address this problem. The first is to simply declare that seeking a wide margin
of victory is a problem for training, not for evaluation. The second is to regard actual tournament results
as an imperfect, ``noisy'' estimate of the outcome in real combat. 
We extend the idea of treating draw as 0.5 to treating near-draw as (e.g.) 0.6 and overwhelming victory as (e.g.) 0.99.

\subsection{Train for Decisive Victory}
The first approach is based on the idea that the margin of victory is important only in that a wide
margin indicates a high probability of future victory -- and that both  Elo and SC-Elo already
properly account to probability of victory.
Thus, the best approach is not to change the rating system but to train the AI to be more aggressive,
without becoming reckless.

The basic approach in reinforcement learning is   to train an agent to take actions which maximize expected 
future rewards from a given situation, based on past experience. The database of past experience
is produced by playing different versions of the agent against itself. Each record is of the form
$(s, a, s', r, z)$ where $s$ is a current state (e.g. state of the Go board), $a$ is action taken (e.g. 
where a stone was placed), $s'$ is the state after the action, $r$ is immediate reward of that action (almost always
zero in Go, but $+1$ if the game ended with a win immediately after that move, $0$ if a tie, and $-1$ if an immediate loss),
and $z$ is the eventual reward at the end of the game (win or lose). Of course, training the agent to play so as to
maximize $z$ is precisely the same as training it to maximize the probability of victory, regardless of the resulting margin
of victory. 

One solution is simply to change the final reward from $z \in \{-1, 0, +1\}$ to something reflecting the margin of victory, while
preserving the  lower bound, zero point, and upper bound. In Go,
the score of each side is the number of points they control at the end of the game.
Because Black moves second,  White has a small advantage, so Black is automatically  awarded a fractional starting score called a 
 {\it{komi}} to reflect that disadvantage. If the  {\it{komi}} is one half point, and the territory on the board
is equally divided at the end, then Black wins by one-half of a stone. For example,
if Black wins by $n$ stones over White, then one could change the final reward from $1$ to the following:
\begin{equation}
z_B = \frac{n+1}{n+2}
\label{eq:Modified_Final_Reward}
\end{equation}

With a fractional {\it{komi}}, ties are impossible so that if Black wins, then White loses: $z_W + z_B = 0$.

Thus, the reward for winning by the slimmest possible margin of a half-stone is $0.6$,
winning by a single stone is $0.667$, and winning by a wider margin produces higher rewards.
Because it is more important to win at all ($+0.6$ from a draw to a minimal victory and $+1.2$ from
minimal loss to minimal victory when ties are impossible) than to win by a huge
margin ($+0.4$ from a minimal victory to an overwhelming victory), the agent does not become reckless. 
Because $0.4$ improvement is not insignificant compared to $0.6$ or $1.2$, the agent still has a strong incentive
to aggressively pursue a wide margin of victory (as long as the probability of victory is not comprised).
The modified reward function can be used both in training and in any associated game play such
as Monte Carlo Tree Search,
neuro-genetic optimization, proximal policy optimization, neuro-fictious self play, and many others.\footnote{The author 
tested a simple MCTS library on both Go and Mancala. Replacing a simple binary reward with
equation \eqref{eq:Modified_Final_Reward} produced noticeably more aggressive play.}

\subsection{Uncertain Victory}
 The second approach reflects the fact that real combat outcomes are stochastic, 
so  a wider ``margin of error'' is better. If Blue just barely beats Red in a strategy game, then this near-tie
outcome  indicates that, given the stochastic nature of combat, that Blue only has a slightly better
than 50\% chance of winning if the game were replayed under the same conditions, except
for different roles of the dice. Similarly, if Blue wins by a wide margin, one should expect
with high probability that Blue would also win again if the game were replayed. 
A second interpretation is entirely analogous to the treatment of ties in  classic Elo:
we simply treat indecisive outcomes as near-ties.
With either interpretation, the key issue is an  expert judgment about the
random variability in a commercial paper strategy game: how big a margin is ``wide'' enough that victory is   decisive?

We will consider the case where player $X$'s score $S_x$ is greater than or equal to $Y$'s score, $S_y$.
To derive a plausible estimator, first we turn to Chebyschev's inequality, which applies to any 
distribution over any random variable $z$ with
finite mean $\mu$ and finite variance $\sigma^2$:

\begin{equation}
P\left[ | z - \mu | \ge k \sigma \right] \le \frac{1}{k^2}, \  \rm{where} \ k \ge 1 
\end{equation}

This implies that ratio of the squared deviation to the variance is the key quantity, for very general distributions.

Next, we observe that if $\Delta  = S_x- S_y$ is a zero margin of victory given scores, then the 
estimated probability of $X$ winning a rematch should be $0.5$ (all other things being equal).
As $X$'s  margin grows, $\Delta  \rightarrow \infty$, then the probability should approach $1$.

These two considerations suggest the following formulation:

\begin{equation}
P\left[ X \succ Y \right] = \frac{\Delta^2 + \delta^2 }{\Delta^2 + 2\delta^2}
\label{eq:Rematch_Victory_Prob}
\end{equation}

The constant $\delta$ represents a subjective judgment as to how wide a margin of victory
is required to make a ``decisive victory'' deserving of full marks, or equivalently, the variance
in scores from the inherently stochastic nature of strategy games. Any margin less than that is scored close to a tie.
Alternatively, equation \eqref{eq:Rematch_Victory_Prob}
codifies a judgment about whether the victory on points in a strategy game 
represents a probable real-world victory or not. In essence, we treat the binary outcome
of the strategy game (on points) as a noisy indicator of the binary outcome of combat.

\begin{table} 
\begin{center}
\begin{tabular} {|c c c | r r |  }
\hline
Blue & Red & Terrain  & $S_B$ & $S_R$  \\
\hline
X & Y & North & 13.7 & 335.2 \\
Y & X & North & 18.6 & 340.5 \\
X & Y & South & 23.2 & 140.7 \\
Y & X & South & 22.1 & 177.5 \\
\hline
\end{tabular}
\caption{Heterogenous, Non-Interactive Competition Scores}
\label{table:HNI_Competition_Scores}
\end{center}
\end{table}

\begin{table} 
\begin{center}
\begin{tabular} {|c c c | r r | c |  }
\hline
Terrain & $X$ Role & $Y$ Role  & $S_X$ & $S_Y$ & Winner \\
\hline
North & Blue & Blue & 13.7 & 18.6 & Y \\
North & Red & Red & 340.5 & 335.2 & X\\
South & Blue & Blue & 23.2 & 22.1 & X\\
South & Red & Red & 177.5 & 140.7 & X\\
\hline
\end{tabular}
\caption{Heterogenous, Non-Interactive Competition Results}
\label{table:HNI_Competition_Results}
\end{center}
\end{table}

To see how this could work in practice, consider an example where agent $X$ and $Y$ get to play either Red or Blue,
in either a Northern terrain or a Southern terrain. The metric for Red is different than that for Blue, and generally 
an order of magnitude higher. Because of the differing terrains, the Blue's scores in the North are generally somewhat 
lower than the South, and vice versa for Red. Purely for illustration,
we suppose that the initial ratings are $R_x = 1250$ and $R_y = 1320$ with $\beta \sigma^2 = 50$ for both.
Illustrative raw scores for  four hypothetical  wargames are presented in Table  \ref{table:HNI_Competition_Scores}. 
Again, because Red and Blue are scored in completely different ways,
one cannot meaningfully compare $13.7$ to $335.2$; it would be like
asking whether 5 kilos was more or less than 7 meters, i.e. whether $(5 \rm{\ kilos}) - (7 \rm{\ meters}) > 0$ or not.
To compare only like-to-like, the data from Table  \ref{table:HNI_Competition_Scores}
is rearranged to produce Table  \ref{table:HNI_Competition_Results}.

\begin{table} 
\begin{center}
\begin{tabular} {|c c c | r r r | c c  |  }
\hline
Terrain & $X$ Role & $Y$ Role  & $S_X$ & $S_Y$ & $\delta$ & Winner & P[Winner] \\
\hline
North & Blue & Blue & 13.7 & 18.6 & 1.6& Y & 0.909 \\
North & Red & Red & 340.5 & 335.2 & 33.8 & X & 0.506 \\
South & Blue & Blue & 23.2 & 22.1 & 2.3 & X & 0.553 \\
South & Red & Red & 177.5 & 140.7 & 16.0 & X & 0.863 \\
\hline
\end{tabular}
\caption{Scoring with Margin of Victory}
\label{table:Noisy_Results}
\end{center}
\end{table}

In the binary SC-Elo formulation, the $A$ term simply counts the number of victories, regardless
of the margin in each. To take into account the margin, we simply redefine $A$ to be the sum over
comparisons of equation \eqref{eq:Rematch_Victory_Prob}. In Chess,
the terms of $A$ can have just three values: $0$ for a loss, $0.5$ for a tie, and $1$ for a win.
We have merely added smooth interpolation so that near-ties are assigned values near $0.5$,
transitioning smoothly to $1$ as the win becomes more significant. The procedure 
is illustrated by 
using the notional data from Table  \ref{table:HNI_Competition_Results} to build
Table \ref{table:Noisy_Results}, assuming that $\delta$ is 10\%
of the RMS scores in each situation.

When we take into account that fact that out of  $X$'s 
three nominal victories only one  was decisive, two were only 
a bit better than ties, and $X$'s one defeat was decisive, then their sum is $A = 2.012$. This corresponds
to a per-game win probability of $0.503$, which is still slightly above the $0.401$ suggested
by the difference in Elo ratings. To illustrate that  Elo always overshoots, we recalculate Table \ref{table:Elo_Overshoot}
with outcomes depending on the margin of victory appears to produce  
Table \ref{table:Elo_with_MV}.
Because the score should be raised,  Elo raises too much, and
eventually to implausibly large values, for reasons discussed earlier.
Again, SC-Elo avoids the overshoot problem.

\begin{table} 
\begin{center}
\begin{tabular} {| r | r r | r |  }
\hline
N & Elo & SC-Elo & $\sigma_{SC}$ \\
\hline
4       & 1270.5 & 1266.0   & 38.6 \\
40     & 1455.0 & 1303.5   & 40.8  \\
400   & 3299.7 & 1319.7   & 16.8  \\
4000 & 21747.3 & 1321.9  & 5.5   \\
\hline
\end{tabular}
\caption{Elo and SC-Elo Scores with Margin of Victory}
\label{table:Elo_with_MV}
\end{center}
\end{table}

\clearpage
\section{Conclusion}

The Self-Consistent Elo (SC-Elo) system presented here is designed to solve five basic problems with the classic Elo rating  system.
These challenges apply to training an AI agent
to play asymmetric games, documenting the AI's skill, and evaluating the components affecting performance.

AI is typically  trained and evaluated on many more games than occur in ordinary Chess tournaments. 
The change in scale causes the classic Elo system to consistently over-correct the new ratings,
potentially by thousands of Elo points. Both the self-consistent posterior maximum likelihood estimator
and the linear least squares estimator eliminate the overshoot. In particular, the least squares
estimator is statistically unbiased.

The various Elo rating systems include an empirical factor, $K$, for the rate of adjustment.
It is based not only on the uncertainty in the ratings of human players or teams but also
on the necessity to avoid systematic drift in the ratings.
While the variability of human performance has been estimated from the extensive literature on ordinary 
Chess tournaments, the variability of each newly trained AI system is completely unknown. 
We present methods to estimate the rating variability based not only on the statistical
uncertainty created by small numbers of games but also on the structural uncertainty
because actual performance might not conform to the fundamental assumptions of the Elo model.
We present several examples of structural uncertainty.
  
The Elo rating system is based on a fairly strong assumption about the ability
to rank players on an ordinal scale. This is known to be inaccurate in important cases,
such as the comparison of players from widely separated eras.
It  is easy to construct small examples which violate this assumption in
much the same way as the Condorcet Voting Paradox. Relaxing
this assumption leads not only to a more complete inventory of the uncertainty in ratings
but also to a simpler and faster algorithm for updating ratings based on linear least squares
estimation of the log-odds of victory. Suggestions for future research to explore
the structure of ratings by means of multi-dimensional clustering are presented.
  
Chess and Go are symmetric games, while commerical paper strategy games typically are not:
opposing players typically
have different forces, on different terrain, and judge 
performance by different criteria,  which  change from scenario to scenario. 
We presented a formulation of the problem which makes it possible to apply
standard ANOVA techniques to assess the imbalance between Red and Blue,
the capability of the agent to play Red or Blue, effects of terrain, and so on.
 
In most strategy games,  succeeding with a comfortable margin is better than 
with a narrow one. In Chess tournaments, outcomes are scored merely 
as win, lose, or tie, with no estimated margin of victory. We present
methods to represent different margins of success similar to the way
a draw is represented as half of a success.

\clearpage 
\appendix
\section{Constant Sum versus Negative Correlation}
\label{sec:CS_vs_NC}

In common usage, the term ``zero sum'' is used fairly loosely to mean something like ``whenever one gains, the other loses''.
This leads to some misunderstanding, because many algorithms depend on meeting the exact zero-sum condition
and hence will fail in cases where the loose condition holds. In particular, algorithms designed for Chess, Go, or Poker
could go awry when applied to an asymmetric competition. It is also possible that the algorithms
might work with little alteration, but they cannot be relied upon until explicitly verified to behave reasonably
in a variable-sum situation.

First, consider a case where Blue's success metric is the exact opposite of Red's. In equation
\eqref{eq:True_Zero_Sum}, each actor's reward depends in some complex way on 
the state state-variables, $x$. Their sum is manifestly zero.
\begin{eqnarray}
S_B & = & +s(x_1, \ldots x_n)  \nonumber \\
S_R & = & -s(x_1, \ldots x_n)
\label{eq:True_Zero_Sum}
\end{eqnarray}

As in all constant-sum situations, the benefit (loss) on Red of changing
any state variable  is exactly opposite to the corresponding loss (benefit) for Blue:

\begin{eqnarray}
\frac{\partial S_B}{\partial x_i} +  \frac{\partial S_R}{\partial x_i}& = &  0
\label{eq:True_Zero_Sum_Slopes}
\end{eqnarray}

Now suppose we change Blue and Red to both be somewhat risk-averse with respect to the same $s$
(like stock brokers who expect
the classic risk-return trade off: junk bonds must pay higher returns to compensate for their higher risk).
In equation \eqref{eq:True_Negative_Correlation}, we simply use one of the standard forms to model
for risk-aversion. As per equation \eqref{eq:True_Zero_Sum}, a
common utility function is equation \eqref{eq:Risk_Averse_Utility} for Blue and the
 opposite in equation \eqref{eq:True_Negative_Correlation} for Red.

\begin{equation}
U_B = \frac{1-e^{-as}}{a} 
\label{eq:Risk_Averse_Utility}
\end{equation}

\begin{equation}
U_R = \frac{1-e^{+as}}{a}
\label{eq:True_Negative_Correlation}
\end{equation}

When $s$ is small, $U_B \approx +s$ and $U_R \approx -s$. However, as $s$ becomes large, $U_B$ approaches
an upper limit of $1/a$, while $U_R$ falls without limit. This is the essence of risk-aversion: the pain of large losses
outweigh the pleasure of equally large gains.

Notice that in equations \eqref{eq:Risk_Averse_Utility} and   \eqref{eq:True_Negative_Correlation}, 
Red and Blue value precisely opposite aspects of the
situation, but their risk-aversion makes the sum variable. In fact, $U_B + U_R$ is zero only when $s=0$, declines
slowly on both side, then accelerating downward rapidly when $1  \ll  |as|$.

Similarly, for large $s$, Blue enjoys few gains from increasing $s$ more (it just approaches $1/a$), while Red suffers large losses.
While the changes always have opposite sign, their sum is never zero (except at the single point $s=0$). When one gains,
the other suffers (which may be the common understanding of ``zero sum''), but not in equal
amounts (the literal requirement of constant-sum).

\begin{equation}
{\rm{sgn}} \left(\frac{\partial U_B}{\partial x_i}\right) + 
{\rm{sgn}} \left( \frac{\partial U_R}{\partial x_i}\right)  =   0  
\end{equation}

\begin{equation}
\frac{\partial U_B}{\partial x_i} +  \frac{\partial U_R}{\partial x_i}  \neq  0 , \ s \neq 0
\label{eq:NonZero_Sum}
\end{equation}

Even though the costs and benefits to Red and Blue are always negatively correlated, and one can gain only when the other loses,
it is not a zero-sum situation. While Red and Blue are obviously in strict competition, algorithms designed for zero-sum
situations would likely yield spurious and unreliable results, or even harmful recommendations.

To borrow an analogy from optimization, the algorithms for convex optimization
(positive curvature) or  linear optimization (zero curvature), typically fail completely for concave optimization
(even slightly negative curvature).
 
\section{Weighted Estimates}
\label{sec:Weighted_Expectations}

Suppose $x_i$ are $N$ estimates of a quantity $X$, but each has
a different level of uncertainty: $x_i  \sim  X \pm \sigma_i$.
We can minimize the expected squared error by choosing $x$ to
minimize the following quantity:

\begin{equation}
S = \sum_{i=1}^N  \left(  \frac{x - x_i}{\sigma_i}  \right)^2
\label{eq:weighted_least_squared_error}
\end{equation}

Taking the derivative with respect to $x$ and setting it to zero, we get the following.

\begin{eqnarray}
0   &=& \sum_{i=1}^N   \frac{x - x_i}{\sigma_i^2}    \nonumber  \\ 
\sum_{i=1}^N   \frac{x  }{\sigma_i^2}   &=& \sum_{i=1}^N   \frac{ x_i}{\sigma_i^2}    \nonumber  \\ 
\end{eqnarray}

This can be simplified by using the ``precision'', which  is defined as the inverse of the variance: $p = 1/\sigma^2$.

\begin{eqnarray}
0   &=& \sum_{i=1}^N p_i \left( x - x_i \right)  \nonumber  \\ 
x   &=&  \frac{ \sum p_i  x_i  }{ \sum p_i }
\end{eqnarray}

The best estimate is just the precision-weighted average of the samples. Notice that if
the uncertainty $\sigma_i$ in one estimate falls toward zero, then the final $x$ 
moves toward $x_i$.

\section{Moving Averages}
\label{sec:moving_averages}

Suppose we have some quantity $x_t$ that is supposed to track some changing goal, $g_t$ over time.
This is similar to 
equation \eqref{eq:Consistent_Elo_Update}, where the rating is adjusted so that the predicted
win probability tracks the changing win frequency over time.
The classic proportional feedback controller adds a correction term each time period
is is a fraction of the error observed that period.

The following demonstrates that this proportional feedback control is equivalent to
a weighted average of the current value and the current goal:

\begin{eqnarray}
x_{t+1} &=&  x_t  + \left( 1 - \lambda \right) \left(  g_t - x_t  \right)   \nonumber  \\
            &=&   \lambda x_t  + \left( 1 - \lambda \right)  g_t   \nonumber  \\
            &=&   \left( 1 - \lambda \right)  g_t   + \lambda x_t   
\label{eq:prop_control_as_average}           
\end{eqnarray}

There is a similar expression for $x_{t}$, $x_{t-1}$, and so on,
so $x_t$ is the exponentially weighted moving average of all the past goals.
If the weight $1 > \lambda > 0 $ changes from period to period, then the past goals will
have weights that decline as a noisy exponential.

\begin{eqnarray}
x_{t+1} &=&  \left( 1 - \lambda \right)  g_t   + \lambda\left[    \left( 1 - \lambda \right)  g_{t-1}   + \lambda x_{t-1}   \right]  \\
            &=&   \left( 1 - \lambda \right) \left[   g_t +  \lambda   g_{t-1} +  \lambda^2 x_{t-1} \right] \nonumber  \\
            & = &  \left( 1 - \lambda \right)  \sum_{i=0}  \lambda^i  g_{t-i}
\label{eq:prop_control_as_long_term_average}           
\end{eqnarray}

This logic can be extended to a state-estimation problem. We will present it as a scalar
problem. Suppose we have a series of states, $x_t$ which may change over time, and a series of noisy
observations of the states, $h_t$. This can be modeled as random changes $\alpha_t$
in the state with
mean zero and variance $\sigma_x^2$ and
random noise in the observation with mean zero and variance $\sigma_h^2$.

\begin{eqnarray}
x_{t+1}  &=&  x_t + \alpha_t  \nonumber \\
h_t  &=&  x_t + \beta_t
\end{eqnarray}

We need to form a series of estimated states, $\hat{x}_t$,  based only on the previous observations $h$.
The sum of all the precision-weighted changes is the following

\begin{equation}
S = \sum_t 
\left(
  \frac{\hat{x}_t  - \hat{x}_{t-1}   }{\sigma_x}
  \right)^2
+
\left(
  \frac{\hat{x}_t  - h_t  }{\sigma_h}
  \right)^2
\end{equation}

The solution is a recursive system of equations, i.e. a precision-weighted moving average:
\begin{equation}
\hat{x}_t = \frac{p_x  \hat{x}_{t-1} + p_h  h_t }{p_x + p_h}
\end{equation}

Notice that while the problem is phrased as an optimization over the whole history,
it can be efficiently implemented by calculating each state only in terms of the most
recent observation and the  immediately
prior state. Each state is a sufficient summary of all the previous observations.
The basic logic of a Kalman filter is analogous, though the matrix algebra is considerably more complex.

If we drop the requirement that the state observation depend only on previous information,
we get the following:

\begin{equation}
\hat{x}_t = \frac{p_x  \left( \frac{\hat{x}_{t+1} + \hat{x}_{t-1} }{2}\right) + p_h  h_t }{p_x + p_h}
\end{equation}

This can be solved efficiently by iterative approximation, where
each new $\hat{x}$ vector is obtained by averaging adjacent
elements of the previous $\hat{x}$ vector with the appropriate $h$. 
 Initial estimates of $\hat{x}_t = h_t$ would probably
converge quickly.

\section{Elo as Moving Averages}

Equation  \eqref{eq:Consistent_Elo_Update} does not strictly match 
equation \eqref{eq:prop_control_as_long_term_average}  , because we are adjusting
$R$ so that the probability derived from $R$ is closer to the observed frequency.
The Elo update has the following form, for a positive constant $c$ and  some monotonically increasing function $f$:

\begin{equation}
x_{t+1} = x_t  + c \left(  f (g_t) - f(x_t)  \right)
\end{equation}

For this case, $f$ is probability as a function of rating, given by
\eqref{eq:Advantage_from_Log_Odds}. 

We will write the series of updates to a rating as $R_t$, the actual number of
wins at time $t$ as $N p_t$, and expected number of wins as $N \hat{p}_t$.
With these substitutions, Equation  \eqref{eq:Consistent_Elo_Update} has the
following form.

\begin{equation}
R_{t+1} = R_t  + NK \left(  p_t -  \hat{p}_t \right)
\end{equation}

 As one can see in 
Figure \ref{fig:elo_vs_english}, the rating-advantage as a function of
probability is fairly linear for reasonably close ratings, because ECF rating
difference is linear in probability for that range. We will take $R_y$ as the
elo-average rating of the opponents in the games being considered.

We can thus write $p_t$ as a linear function of the ideal rating,
which is the goal $G_t$ we would like $R$ to approach.
Just as in equation \eqref{eq:scelo_large_limit}, $G_t$
is precisely the value  needed to  reproduce exactly the observed win-fraction.
Similarly,
$ \hat{p}_t$ is that same linear function of $R_{t+1}$ for SC-Elo:

\begin{eqnarray}
p_t &=&   a (G_t  - R_y) + b   \nonumber   \\
\hat{p}_{t}  &=&   a (R_{t+1} - R_y ) + b 
\end{eqnarray}

where  $a = \partial p  /  \partial  R$.

\begin{eqnarray}
R_{t+1}  &=& R_t  + NK \left( \left[ a (G_t  - R_y) + b \right]  -  \left[ a (R_{t+1} - R_y ) + b \right] \right)  \nonumber   \\
R_{t+1}  &=& R_t  +a  NK \left( G_t  - R_{t+1}  \right)  \nonumber   \\
R_{t+1}  &=& \lambda  R_t  +  \left( 1- \lambda  \right)  G_t
\label{eq:SC-Elo-as-moving-average}
\end{eqnarray}

where $\lambda = 1/(1+aNK)$. As expected, smaller $K$ values cause less adjustment and 
larger $K$ values cause more.
Because the $aN$ will vary, the past goals
are weighted with a noisy exponential.
Note that if $N$ increases without limit,
equation \eqref{eq:SC-Elo-as-moving-average} will simply bring $R_{t+1}$ 
closer and closer to  $G_t$.
This is the same limit as  equation \eqref{eq:scelo_large_limit}  in section
\ref{sec:Large_Tournament_Limit}.

A very similar derivation can be done with classic Elo updates to
get the following formula.

\begin{eqnarray}
R_{t+1} &=&  \left(  1 - aNK \right) R_t  + aNK G_t  \nonumber   \\ 
             &=&   R_t   +  aNK  \left( G_t - R_t  \right) 
\label{eq:Classic-Elo-as-moving-average}
\end{eqnarray}

 Note that if $N$ increases without limit, 
 the right hand side of equation \eqref{eq:Classic-Elo-as-moving-average} will 
also grow without limit, which produces the overshoot behavior  discussed for
equation \eqref{eq:Classic_Elo_Update_N}.

We have thus reached the form of equation  \ref{eq:prop_control_as_average}: both classic Elo and SC-Elo are
actually moving averages of the ``ideal rating''.

\section{Simple Mean and Variance}

\begin{table} 
\begin{center}
\begin{tabular} {| r  r | r r | r r |  }
\hline
$W_x$ & $L_x$  & $\mu^N$ & $\sigma^N$ & $\mu^A$ & $\sigma^A$  \\
\hline
    1 &     0 &   172.5 & 258.2 & 169.0 & 217.3 \\
    1 &     1 &   0.0 & 197.4 & 0.0 & 167.2 \\

\hline
    2 &     0 &   258.7 & 241.3 & 264.4 & 225.1 \\
    2 &     1 &    86.9 & 177.3 & 85.2 & 155.6 \\
    2 &     2 &   0.0 & 154.4 & 0.0 & 138.2 \\
\hline
    5 &     0 &   393.0 & 222.3 & 429.8 & 254.0 \\
    5 &     1 &  223.1 & 158.2 & 228.8 &  148.2  \\
    5 &     2 &  136.1 & 131.9 & 135.3 & 123.1 \\
    5 &     3 &  78.2& 118.5 &  77.7 &  111.0 \\
    5 &     5 &  0.0 & 104.6 &  0.0 &  99.0 \\

\hline
\end{tabular}
\caption{Approximation mean and standard Deviation, small numbers}
\label{table:approx_mean_stdv_small}
\end{center}
\end{table}

For any random variable, $x$, we can define the following quantities, when the expectations
of the first and second power actually exist.

\begin{eqnarray}
\mu_x &=& E(x)  \nonumber \\
\sigma^2_x  &=&  E([x-\mu]^2)  \nonumber \\
                     &=& E(x^2) - E(x)^2
\end{eqnarray}

In particular, consider a simple random variable which has 50/50 odds of being
either $x_0 = m-s$ or $x_1 = m+s$. In this case, $\mu = m$ and $\sigma = s$.

\begin{eqnarray}
\mu   &=&  m \nonumber   \\
         &=& \frac{x_1 + x_0}{2} 
\end{eqnarray}

\begin{eqnarray}
\sigma   &=&  s  \nonumber   \\
         &=& \frac{x_1 - x_0}{2}
\end{eqnarray}

\begin{table} 
\begin{center}
\begin{tabular} {| r  r | r r | r r |  }
\hline
$W_x$ & $L_x$  & $\mu^N$ & $\sigma^N$ & $\mu^A$ & $\sigma^A$  \\
\hline
    40 &   10 &  234.4 & 60.1 & 234.4 & 59.6  \\
    30 &     20 & 69.9 & 49.6 & 69.0 & 49.0 \\
    25 &     25 &  0.0 & 48.6 &  0.0 &  48.0 \\

\hline
    70 &     30 & 145.5 & 37.6 & 145.5 & 37.4  \\
    50 &     50 &  0.0 & 34.6 &  0.0 &  34.3 \\

\hline
150 &   50 &    189.7 &  28.3 &  189.7 &  28.2 \\
100 & 100 &     0.0 &   24.5 &  0.0   &   24.4 \\

\hline
\end{tabular}
\caption{Approximation mean and standard Deviation, large numbers}
\label{table:approx_mean_stdv_large}
\end{center}
\end{table}

We can use these relations to approximate the mean and standard deviation of $A_{xy}$.
The posterior estimates for the probability that X defeats Y
are given by equation \eqref{eq:Binomial_Parameters}
where $N = W_x = L_y$ and $M = L_x = W_y$.

First, we get the one-sigma ranges on the posterior probability  from equation \eqref{eq:Binomial_Parameters}:
\begin{eqnarray}
x_0   &=&  \mu - \sigma  \nonumber  \\
x_1   &=&  \mu + \sigma  
\end{eqnarray}

We get the corresponding $A$ values:
\begin{eqnarray}
A_0  &=&  \frac{1}{\beta} \ln\left( \frac{x_0}{1 - x_0} \right)   \nonumber  \\
A_1  &=&  \frac{1}{\beta} \ln\left( \frac{x_1}{1 - x_1} \right)  
\end{eqnarray}

Finally, we define our approximations for the mean and standard deviation:

\begin{eqnarray}
\mu^A_{xy} &=&  \frac{A_1 + A_0}{2}   \nonumber  \\
\sigma^A_{xy} &=&  \frac{A_1 - A_0}{2} 
\end{eqnarray}

These approximations can easily be checked against
$\mu^N$ and $\sigma^N$ as calculated by numerical integration;
see tables \ref{table:approx_mean_stdv_small}
and  \ref{table:approx_mean_stdv_large}.
The standard deviations  are of approximately the same order of magnitude
as Elo's results in Figure \ref{fig:elo-reliability}.
Even for the small sizes, the difference is means is much less than
than the uncertainty in either.

\section{Risk-Averse Betting}
\label{sec:Betting}

We consider the situation of betting on a number of exclusive and exhaustive outcomes. It might be eight chess challengers
who want to play the champion, six horses running a race, or a hundred football teams,
only one of which can eventually win the world title. ``The House'' would like not only to offer as much opportunity to
bettors as possible  but also to get a guaranteed percentage, no matter which option occurs.

Clearly, we can not play the odds.
First, whose odds should we use? Any particular set
is just subjective
Second, even if "true probabilities" and were known to us,
unlikely events do happen. We might always win in the
long run, but we could lose big in the short term when
long-shots win.

\subsection{Individual Betting}

First we will examine the behavior of bettors, as the design of a betting system must work with that behavior.
Let us suppose that there are $N$ exclusive and exhaustive outcomes. If someone bets $b$ on outcome $i$, they lose that amount
when placing the bet. If they win, then they get back $R = rb$, for a net gain of $b(r-1)$. Note that $r > 1$ always,
because $r < 1$ would mean that the bettor lost money by winning.

The risk-neutral criterion for bet $i$ to be worthwhile is the first line of equation (\ref{eq:Risk_Neutral_Criterion}).
The second two lines are equivalent and will appear at various points later in this paper.

\begin{eqnarray}
p_ib(r_i - 1) + (1-p_i)(-b)  &\ge& 0 \nonumber \\
\frac{p(r-1)}{1-p}            &\ge& 1 \nonumber \\
p_i r_i                             &\ge& 1 
\label{eq:Risk_Neutral_Criterion}
\end{eqnarray}

The bettor sees it as an exactly fair bet when equality holds;
a product greater than one would have positive expected value.
The following three forms of the equality will be used repeatedly below:

\begin{eqnarray}
p_i r_i &=& 1  \nonumber \\
p_i &=& \frac{1}{r_i}  \nonumber \\
r_i &=& \frac{1}{p_i}
 \label{eq:Fair_Odds}
\end{eqnarray}

 Unfortunately,
risk neutrality is unrealistic in that a truly risk-neutral bettor would make vast bets for only slight expected returns.
Of course, few people are risk-neutral, which is why financial markets demand higher returns to compensate for higher risk: 
they are risk averse, and the  higher the risk is the  higher the compensation  required to overcome that aversion.

As mentioned above, a common and well-founded model of utility is the following:

\begin{equation}
u(x)  = \frac{1-e^{-ax}}{a} 
\label{eq:IndividualUtilities}
\end{equation}

As before when $|ax| \ll 1$, $u(x) \approx x$. As $x$ grows without limit, the utility asymptotically approaches $1/a$,
while the dis-utility of losses worsens without bounds.

The parameter $a$ controls the shape, and it fortunately has an intuitive interpretation. If losing $D$ of currency hurts twice
as much as gaining $D$ helps, then $a = \ln(2)/D$.

With this utility function, one can work out the optimal bet, given $p$, $r$, and $a$ as above, simply by
setting the derivative to zero and solving:

\begin{equation}
arb = \ln \left[ \frac{p(r-1)}{1-p} \right]   
\label{eq:OptimalBetSize}
\end{equation}

Substituting the expression for $a$ in terms of $D$, we get the following:

\begin{equation}
b  =   \frac{D}{\ln(2) r}    \ln \left[ \frac{p(r-1)}{1-p} \right] 
\label{eq:Betsize_from_LogOdds_one}
\end{equation}

Note that $b \ge 0$ exactly when the second line of equation (\ref{eq:Risk_Neutral_Criterion})  holds. Thus,
the decision whether or not to bet is made in a risk-neutral way, but the size of the bet is determined 
by risk-attitude. Risk-averse actors make small bets and risk-tolerant actors make large bets, so the
market as a whole acts risk-neutral, even though every individual actor may be risk-averse. Their
endowment-weighted risk attitudes determine the volume of trading.

We can slightly rearrange equation \eqref{eq:Betsize_from_LogOdds_one}
to emphasize that the probability-related component is the log-odds:

\begin{equation}
b  =   \frac{D}{\ln(2) r} \left(   \ln \left[ r-1 \right]  +    \ln \left[ \frac{p }{1-p} \right] \right)
\label{eq:Betsize_from_LogOdds_two}
\end{equation}

Estimating the log-odds as accurately as possible is exactly what equation
\eqref{eq:Nonadditive-Error-Minimization} does.

\clearpage
\bibliography{elo-large-tournaments}

\begin{thebibliography}{10}

\bibitem{BFM76}
Frank Chadwick.
\newblock Battle for {M}oscow, 1988.

\bibitem{Condorcet}
Marquis de~Condorcet Marie Jean Antoine Nicolas~de Caritat.
\newblock Essay on the application of analysis to the probability of majority
  decisions.
\newblock 1785.

\bibitem{PGG76}
James Dunnigan.
\newblock Panzergruppe {G}uderian, 1976.

\bibitem{efird2015toward}
Brian Efird and Ben Wise.
\newblock Toward the integration of policymaking models and economic models.
\newblock In {\em Presented at the 18th Annual Conference on Global Economic
  Analysis, Melbourne, Australia}, Department of Agricultural Economics, Purdue
  University, West Lafayette, IN, 2015.

\bibitem{RatingChessPlayers}
Arpad Elo.
\newblock {\em Rating Chess Players, Past and Present}.
\newblock Arco Publishing, New York, New York, 1986.

\bibitem{Kiraly_ModellingCompetitiveSports}
Franz Kiraly and Zhaozhi Qian.
\newblock Modelling competitive sports: {B}radley-{T}erry-{E}lo models for
  supervised and on-line learning of paired competition outcomes, 2017.

\bibitem{RankingSystemsInFootball}
Jan Lasek, Zoltan Szlavik, and Sandjai Bhulai.
\newblock The predictive power of ranking systems in association football.
\newblock {\em International Journal of Applied Pattern Recognition},
  1(1):27--46, 2013.

\bibitem{Massey_StatModelsRanking}
Kenneth Massey.
\newblock Statistical models applied to the rating of sports teams.
\newblock Master's thesis, Bluefield College, Bluefield, VA, 1997.

\bibitem{Napoli2018}
Christopher Napoli, Ben Wise, David Wogan, and Lama Yaseen.
\newblock {\em Policy Options for Reducing Water for Agriculture in Saudi
  Arabia}, pages 211--230.
\newblock Springer Singapore, Singapore, 2018.

\bibitem{MasteringGo2017}
David Silver, Julian Schrittwieser, Simonyan Karen, Ioannis Antonoglou, Aja
  Huang, Arthur Guez, Thomas Hubert, Lucas Baker, Matthew Lai, Adrian Bolton,
  Yutian Chen, Timothy Lillicrap, Hui Fan, Laurent Sifre, George van~den
  Driessche, Thore Graepel, and Demis Hassabis.
\newblock Mastering the game of {Go} without human knowledge.
\newblock {\em Nature}, 39:354--359, 10 2017.

\bibitem{Sublinear16}
Ben Wise.
\newblock Group choice with interdependent sublinear voting.
\newblock Technical Report KS-1629-MP023B, King Abdullah Petroleum Studies and
  Research Center, Riyadh, KSA, 2016.

\bibitem{IntroKTAB15}
Ben Wise, Leo Lester, and Brian Efird.
\newblock An introduction to the {KAPSARC} toolkit for behavioral analysis
  ({KTAB}) using one-dimensional spatial models.
\newblock Technical Report KS-1517-DP011A, King Abdullah Petroleum Studies and
  Research Center, Riyadh, KSA, 2015.

\bibitem{MultiDimKTAB15}
Ben Wise, Leo Lester, and Brian Efird.
\newblock Multidimensional bargaining using {KTAB}.
\newblock Technical Report KS-1524-DP018A, King Abdullah Petroleum Studies and
  Research Center, Riyadh, KSA, 2015.

\end{thebibliography}

\end{document}